\documentclass[lettersize,journal]{IEEEtran}

\UseRawInputEncoding
\usepackage{amsmath,amsfonts}
\usepackage{algorithmic}
\usepackage{algorithm}
\usepackage{array}
\usepackage[caption=false,font=normalsize,labelfont=sf,textfont=sf]{subfig}
\usepackage{textcomp}
\usepackage{stfloats}
\usepackage{url}
\usepackage{verbatim}
\usepackage{graphicx}
\usepackage{caption}
\usepackage{cite}
\usepackage{xspace}
\usepackage{booktabs} 
\usepackage{siunitx}
\usepackage{orcidlink}
\usepackage{pbalance}
\usepackage{todonotes}
\usepackage{paralist}
\usepackage{lipsum}
\usepackage{makecell} 
\usepackage{arydshln} 
\usepackage{acro} 
\DeclareAcronym{bcs}{
    short = BCS,
    long  = Base Control Software
}
\DeclareAcronym{cav}{
    short = CAV,
    long  = Connected and Automated Vehicle,
}
\DeclareAcronym{dds}{
    short = DDS,
    long  = Data Distribution Service
}
\DeclareAcronym{cnn}{
    short = CNN,
    long  = Convolutional Neural Network
}
\DeclareAcronym{ros}{
    short = ROS,
    long  = Robot Operating System
}
\DeclareAcronym{rs}{
    short = RS,
    long  = Robot Swarm,
}

\hypersetup{
    colorlinks,
    linkcolor={black},
    citecolor={black},
    urlcolor={black}
}

\newcommand{\TestbedCount}{23 }
\newcommand{\TestbedCountRS}{seven }

\newcommand{\SingleAgentCount}{four }

\newcommand{\SingleAgentCountRS}{three }
\newcommand{\mainTestbeds}{nine }

\newcommand{\characteristicsCount}{62 }
\newcommand{\domainName}{CAV/RS}
\newcommand{\weburl}{https://bassamlab.github.io/testbeds-survey }

\makeatletter
\def\ps@IEEEtitlepagestyle{%
  \def\@oddfoot{\mycopyrightnotice}%
  \def\@evenfoot{}%
}
\def\mycopyrightnotice{%
	\begin{minipage}{\textwidth}
		\centering \scriptsize
        \copyright \the\year{} IEEE. Personal use of this material is permitted. Permission from IEEE must be obtained for all other uses, in any current or future media, including reprinting/republishing this material for advertising or promotional purposes, creating new collective works, for resale or redistribution to servers or lists, or reuse of any copyrighted component of this work in other works.\hfill
	\end{minipage}
	\gdef\mycopyrightnotice{}
}
\makeatother

\begin{document}

\title{A Survey on Small-Scale Testbeds for Connected and Automated Vehicles and Robot Swarms}

\author{
    Armin Mokhtarian$^{1}$\,\orcidlink{0000-0002-5345-4538},
    Jianye Xu$^{1}$\,\orcidlink{0009-0001-0150-2147},
    Patrick Scheffe$^{1}$\,\orcidlink{0000-0002-2707-198X},
    Maximilian Kloock$^{1}$\,\orcidlink{0000-0002-3470-6429},
    Simon  Sch\"afer$^{1}$\,\orcidlink{0000-0002-6482-2383},
    Heeseung Bang$^{2}$\,\orcidlink{0000-0003-4331-1952},
    Viet-Anh Le$^{2}$\,\orcidlink{0000-0002-9829-7150},
    Sangeet Ulhas$^{3}$ \orcidlink{0009-0006-1292-8959},
    ~\IEEEmembership{Student~Members,~IEEE}, 

    Johannes Betz$^{4}$ \orcidlink{0000-0001-9197-2849},
    Sean Wilson$^{5}$ \orcidlink{0000-0002-6282-4772},
    Spring Berman$^{3}$\,\orcidlink{0000-0001-9239-0509},
    Liam Paull$^{6}$ \orcidlink{0000-0003-2492-6660},
    ~\IEEEmembership{Members,~IEEE},

    Amanda Prorok$^{7}$\orcidlink{0000-0001-7313-5983},
    ~Bassam Alrifaee$^{8}$\orcidlink{0000-0002-5982-021X},
    ~\IEEEmembership{Senior Members,~IEEE}.
	\thanks{
        $^{1}$Armin Mokhtarian, Jianye Xu, and Patrick Scheffe are Ph.D. students and Maximilian Kloock is a former Ph.D. student with the Department of Computer Science, RWTH Aachen University, Germany.
        {\tt\small \{lastname\}@embedded.rwth-aachen.de}.
        Testbed affiliation: CPM Lab.
    }
	\thanks{
        $^{2}$Heeseung Bang is a postdoctoral researcher and Viet-Anh Le is a Ph.D. student with the Department of Mechanical Engineering, University of Delaware, USA.
        {\tt\small \{heeseung,vietale\}@udel.edu}.
        Testbed affiliation: IDS3C.
    }
    \thanks{
        $^{3}$Sangeet Ulhas is a Ph.D. student and Spring Berman is an associate professor with the Department of Mechanical and Aerospace Engineering, Arizona State University, USA.
        {\tt\small \{sulhas, spring.berman\}@asu.edu}.
        Testbed affiliation: co-creators of CHARTOPOLIS. 
    }
    \thanks{
        $^{4}$Johannes Betz is an assistant professor with the TUM School of Engineering and Design, Technical University Munich, Germany.
        {\tt\small johannes.betz@tum.de}.
        Testbed affiliation: F1TENTH.
    }
    \thanks{
        $^{5}$Sean Wilson is a senior research engineer with the School of Electrical and Computer Engineering, Georgia Institute of Technology, USA.
        {\tt\small sean.wilson@gtri.gatech.edu}.
        Testbed affiliation: Robotarium.
    }
    \thanks{
        $^{6}$Liam Paull is an associate professor with the Department of Computer Science and Operations Research, Universit\'e de Montr\'eal, Canada.
        {\tt\small paulll@iro.umontreal.ca}.
        Testbed affiliation: co-creator of Duckietown.
    }
    \thanks{
        $^{7}$Amanda Prorok is a professor with the Department of Computer Science and Technology, University of Cambridge, UK.
        {\tt\small asp45@cam.ac.uk}.
        Testbed affiliation: creator of Cambridge Minicar.
    }
    \thanks{
        $^{8}$Bassam Alrifaee is a professor with the Department of Aerospace Engineering, University of the Bundeswehr Munich, Germany.
        {\tt\small bassam.alrifaee@unibw.de}.
        Testbed affiliation: creator of CPM Lab.
        \\
    }
    %

    \thanks{We acknowledge the financial support for this project by the Collaborative Research Center / Transregio 339 of the German Research Foundation (DFG).}

}


\IEEEpubid{0000--0000/00\$00.00~\copyright~2021 IEEE}

\maketitle

\begin{abstract}
Connected and automated vehicles and robot swarms hold transformative potential for enhancing safety, efficiency, and sustainability in the transportation and manufacturing sectors. 
Extensive testing and validation of these technologies is crucial for their deployment in the real world.
While simulations are essential for initial testing, they often have limitations in capturing the complex dynamics of real-world interactions. This limitation underscores the importance of small-scale testbeds.
These testbeds provide a realistic, cost-effective, and controlled environment for testing and validating algorithms, acting as an essential intermediary between simulation and full-scale experiments. 

This work serves to facilitate researchers' efforts in identifying existing small-scale testbeds suitable for their experiments and provide insights for those who want to build their own.
In addition, it delivers a comprehensive survey of the current landscape of these testbeds.
We derive \characteristicsCount characteristics of testbeds based on the well-known {\it sense-plan-act} paradigm and offer an online table comparing \TestbedCount small-scale testbeds based on these characteristics.
The online table is hosted on our designated public webpage \url{\weburl}, and we invite testbed creators and developers to contribute to it.
We closely examine nine testbeds in this paper, demonstrating how the derived characteristics can be used to present testbeds.
Furthermore, we discuss three ongoing challenges concerning small-scale testbeds that we identified, i.e., small-scale to full-scale transition, sustainability, and power and resource management.
\end{abstract}


\begin{figure*} [ht!]
\includegraphics[width=\textwidth]{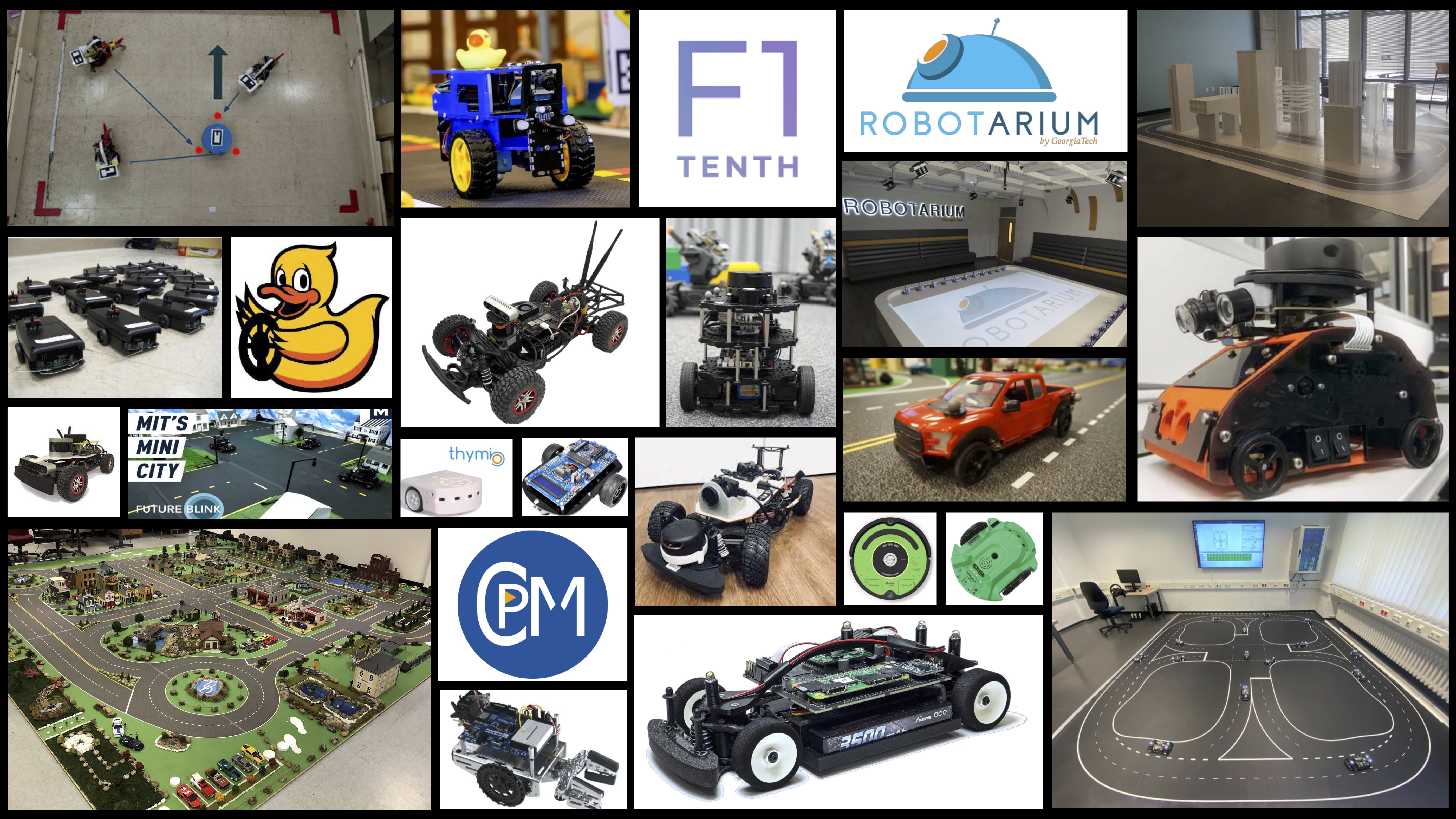}
\captionof{figure}{Collage showcasing diverse testbeds in the realm of Connected and Automated Vehicles and Robot Swarms.}\label{cola}
\end{figure*}

\maketitle
\thispagestyle{IEEEtitlepagestyle}
\pagestyle{plain}


\section{Introduction}
\Acp{cav} and \acp{rs} can significantly improve the safety, efficiency, and sustainability of the transportation and manufacturing sectors. Yet, the deployment and integration of these technologies necessitate rigorous testing and validation to ensure they perform well under varied real-world conditions.
Although simulations facilitate rapid testing and validation, they suffer from limitations in replicating complex real-world conditions and uncertainties, rendering the results possibly unreliable.
In contrast, full-scale experiments, though providing trustworthy results, can be prohibitively expensive, time-consuming, and difficult to reproduce.
Small-scale testbeds aim to bridge the gap between simulations and full-scale experiments by providing cost-effective and controlled environments. These testbeds can simulate real-world scenarios with varying degrees of complexity and realism for testing and validating algorithms under different conditions. 

Throughout this paper, we will use the term {\it testbed} to refer to small-scale testbeds and the term {\it \domainName{}} to denote the domain of \acp{cav} and \acp{rs}.
In addition, we will use the term {\it agent} to refer to a vehicle in \acp{cav} or a robot in \acp{rs}.

Surveys of small-scale testbeds for \domainName{} are essential for helping researchers understand the current landscape of technologies and methodologies in this domain. These surveys facilitate the sharing of knowledge and experiences, enabling researchers to build upon existing foundations and avoid redundant efforts. However, existing survey papers often lack comprehensiveness regarding the characteristics of testbeds, primarily focusing on research areas and only certain characteristics. For example, while \cite{babu2022comprehensive} provides a comprehensive overview of simulators, emulators, and testbeds for vehicular adhoc networks, it lacks granular details on specific characteristics of testbeds, such as those related to physical properties and hardware specifications.
Moreover, they may capture a snapshot of the current landscape of existing testbeds at the time, but the state of the art is continually evolving.
This deficiency hinders researchers' abilities to make informed decisions about which testbeds might best meet their specific needs and objectives, or gain insights for building their own testbeds.

Recognizing these limitations, we identify a need for a new survey paper that comprehensively captures the latest developments in small-scale testbeds and has a designated website to update outdated information. This paper aims to offer a structured and up-to-date overview, detailing the capabilities of various testbeds and assisting researchers in selecting the most appropriate ones for their experiments, or conveying insights for them to build their own testbeds. 

Advancements in the \domainName{} domain are facilitated by testbeds that focus on particular use cases. Each testbed must meet distinct requirements to support the development and testing of these use cases effectively. Adopting the well-known sense-plan-act paradigm \cite{brooks1986robust}, we divide the functional components of these testbeds into sensing, planning, and acting, each necessitating specific testing needs. Further, we derive characteristics to describe whether and how a specific requirement is fulfilled. The significance of certain characteristics varies depending on the researcher's specific focus. 
For instance, when considering sensing, information about the available sensors is important, as it describes the hardware available for tasks like localization and object identification.
In contrast, for a researcher concentrating on planning, information on sensors may be irrelevant, as long as the current location of an agent is accessible. In this case, instead of \textit{sensors}, a characteristic such as \textit{agent count} would serve as a more important factor, as it impacts the complexity and scalability of planning algorithms.
In terms of acting, characteristics like \textit{agent count} are not as important as actuator hardware, which directly influences the precision and range of actions an agent can perform, thereby impacting the effectiveness of its response to the planned strategies in dynamic environments.

\subsection{Contributions} \label{sec:contributions}
\begin{figure}[t!]
\includegraphics[width=0.48\textwidth]{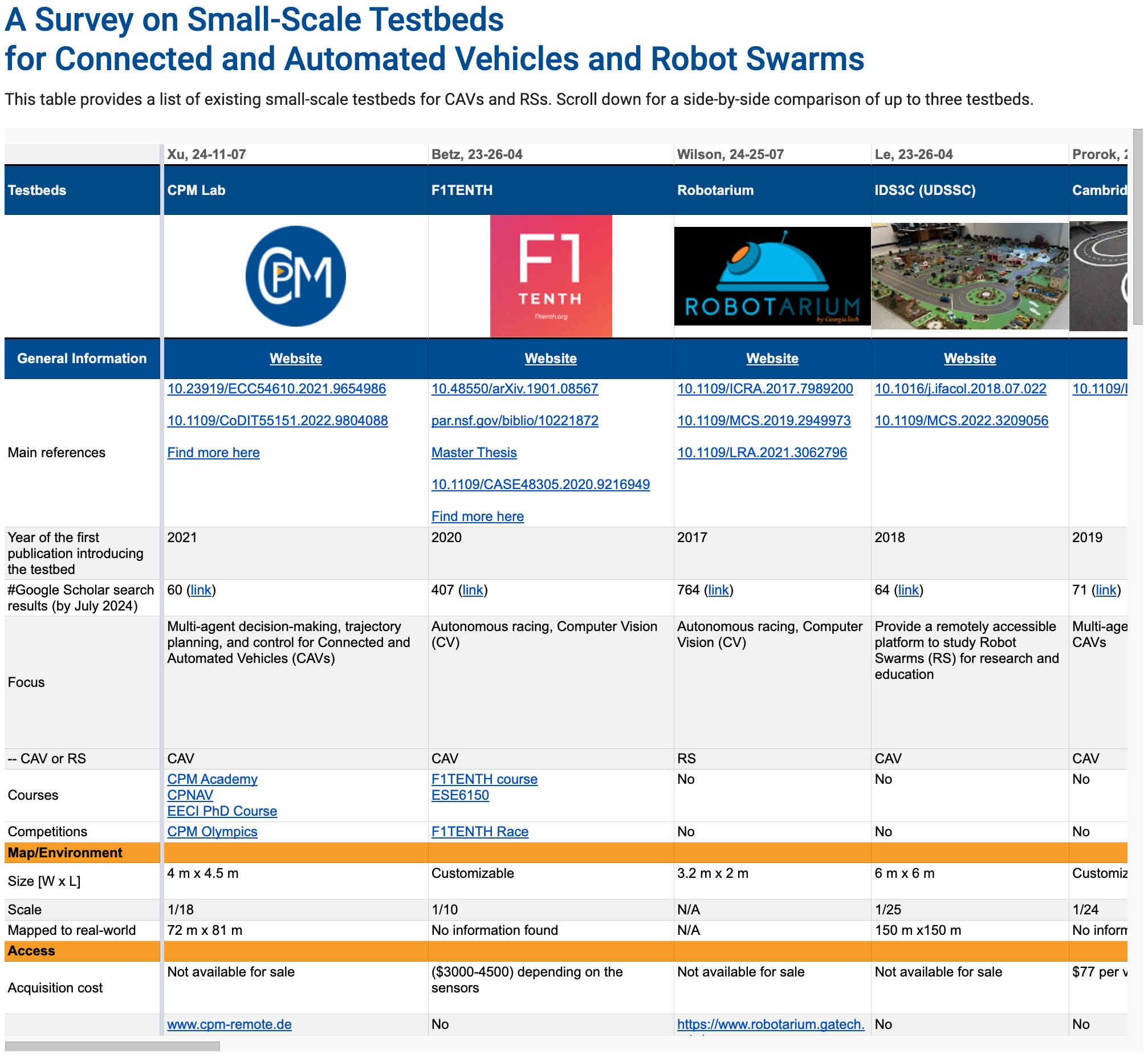}
\caption{A screenshot of the publicly accessible online table \cite{webpage} that lists all the testbeds investigated in this study.}\label{screenshot}
\end{figure}

This work serves to aid researchers, regardless of their experience level in \domainName, in quickly identifying and selecting suitable testbeds that align with their experimental requirements. It also serves as a resource for those considering building new testbeds by highlighting essential software and hardware considerations and ongoing challenges with small-scale testbeds. The contribution of this work is fourfold.
\begin{itemize}
    \item It derives \characteristicsCount characteristics of small-scale testbeds based on the sense-plan-act paradigm.
    \item It introduces a continuously updated online table on a dedicated webpage\footnote{\url{\weburl}} \cite{webpage} that compares \TestbedCount existing testbeds \cite{kloock_cyber-physical_2021, okelly_f1tenth_2020, pickem2017robotarium, stager_scaled_2018, hyldmar_fleet_2019, paull_duckietown_2017, ulhas_chartopolis_2022, reina2017ARK, antoun2016kilogrid, blumenkamp2024cambridge, kannapiran_go-chart_2020, carron_chronos_2023, wilson_pheeno_2016, buckman_evaluating_2022, schwab_experimental_2020, graham_abstractions_2009, dong_mixed_2023, crenshaw_tanya_l_and_beyer_steven_upbot_2010, liniger_optimization-based_2015, tiedemann_miniature_2022, tian2024icat, michael2008experimental, wang2024introduction} using the derived characteristics.
    \item It details these characteristics through an in-depth exploration of \mainTestbeds selected testbeds \cite{kloock_cyber-physical_2021, okelly_f1tenth_2020, pickem2017robotarium, stager_scaled_2018, hyldmar_fleet_2019, paull_duckietown_2017, ulhas_chartopolis_2022, reina2017ARK, antoun2016kilogrid}.
    \item It discusses three identified ongoing challenges with small-scale testbeds, i.e., small-scale to full-scale transition, sustainability, and power and resource management, each illustrated with specific examples from the selected testbeds.
\end{itemize}
This online table \cite{webpage} allows ongoing contributions that keep the information on the testbeds up to date to benefit the research community working with small-scale testbeds for \domainName.
In addition, we invite testbed creators and developers to contribute to the online table. Furthermore, the online table additionally includes and compares \SingleAgentCountRS commercially available robots \cite{mondada1994mobile, mondada2009puck, rubenstein2012kilobot} that are widely adopted in building \ac{rs} testbeds.
Figure \ref{cola} displays a collage of some testbeds included in the online table.
Figure \ref{screenshot} shows a screenshot of the webpage hosting the online table. 

\subsection{Related Work}
\label{subsec:relatedWork}
A recent review by Caleffi et al. \cite{caleffi_systematic_2023_1} summarizes existing literature on \ac{cav} testbeds. While it identifies trends, such as the recent increase in publications and the predominant focus on software, it does not conduct a comparative analysis. Despite various surveys that have categorized testbeds, a unified and detailed comparative analysis remains absent, particularly for recent developments. The most recent comparative paper dates back to 2013 \cite{jimenez-gonzalez_testbeds_2013}, in which the authors focus on \acp{rs} and mainly describe two approaches to classify the testbeds. 
The first approach is a classification by level of complexity, distinguishing between 
\begin{inparaenum}[(i)]
\item ``non-integrated,'' e.g., multi-robot testbeds, 
\item ``partially integrated,'' e.g., multi-robot testbeds with sensor networks, and 
\item ``highly integrated,'' e.g., federated testbeds.
\end{inparaenum}
In this context, {\it integrated} refers to the level of coordination and combination among different components or subsystems within the testbed environment.
The second approach is a classification according to 
    \begin{inparaenum}[(i)]
        \item range of supported experiments (application-driven, functionality-driven and general-purpose),
        \item architecture flexibility,
        \item target users of the testbed,
        \item proximity between testbed experiments and the final application, and
        \item use of the testbed in real deployments (such as industrial sites and urban environments).
    \end{inparaenum}
However, both approaches describe testbeds on a rather abstract level and do not adequately describe detailed characteristics.

Recent studies have started to address these shortcomings by surveying the latest developments of small-scale agents.
For instance, \cite{li2024towards} provides a comprehensive survey on recent developments in autonomous driving with small-scale cars, covering aspects such as widely used small-scale testbeds for \acp{cav}, autonomous-driving tasks (e.g., localization, lane keeping, and collision avoidance), and commonly used sensors with their application scenarios.
We have noticed that recent papers discussing small-scale testbeds often feature tables in their related work section that compare testbeds. However, these tables vary significantly in both covered testbeds and the level of detail provided. 
For example, the tables in both \cite{li2024towards} and \cite{samak_autodrive_2023} contain information about sensors, computation units, agent dynamics, and simulation platforms. 
While the table in \cite{samak_autodrive_2023} additionally provides details on cost, V2X support, and API support, the table in \cite{li2024towards} includes information on the agent's size, the software platform, and whether the testbed is open-source or commercial.

To address the redundancy observed in the literature and unify the fragmented information available, we aim to provide a relatively comprehensive, up-to-date online table that compares existing testbeds for \domainName. To maintain the relevance and accuracy of the information, we are committed to regularly updating our online table \cite{webpage}. The online table currently contains \TestbedCount different testbeds.
For the purposes of our discussion in this paper, we selected \mainTestbeds testbeds to focus on in more detail: 
The Cyber-Physical Mobility Lab (\mbox{CPM Lab}) \cite{kloock_cyber-physical_2021}, 
F1TENTH \cite{okelly_f1tenth_2020}, 
the Robotarium \cite{pickem2017robotarium}, 
IDS3C \cite{stager_scaled_2018}, 
the Cambridge Minicar~\cite{hyldmar_fleet_2019}, 
Duckietown \cite{paull_duckietown_2017}, 
\mbox{CHARTOPOLIS} \cite{ulhas_chartopolis_2022}, 
Augmented Reality for Kilobots (ARK) \cite{reina2017ARK}, and
Kilogrid \cite{antoun2016kilogrid}.
The selection of these testbeds is based on two factors.
Firstly, the co-authors possess substantial familiarity with these testbeds, given their roles as developers or creators of these testbeds.
Secondly, these testbeds exhibit diverse characteristics, which provide varied insights into the sense-plan-act paradigm.
This diversity is crucial for a comprehensive analysis of the paradigm across different contexts and applications.
Note that we consider the online table a living part of this paper. Therefore, the absence of other testbeds in the  printed or digital version of the paper does not suggest they are of lesser importance than the selected ones. Rather, the chosen testbeds are well-suited to illustrate the specific points and analyses discussed in this paper.

Among the \TestbedCount selected testbeds, only \TestbedCountRS of them are \ac{rs} testbeds, i.e., \cite{pickem2017robotarium, reina2017ARK, antoun2016kilogrid, wilson_pheeno_2016, schwab_experimental_2020, crenshaw_tanya_l_and_beyer_steven_upbot_2010, michael2008experimental}.
To balance this disproportion, we add \SingleAgentCountRS swarm robots to our online table \cite{webpage}.
Although they are not testbeds, they have already been widely used to develop \ac{rs} testbeds and hold great potential to develop new \ac{rs} testbeds.
These robots are
Khepera \cite{mondada1994mobile, soares2016khepera}, 
e-Puck \cite{mondada2009puck, cianci2007communication}, and
Kilobot \cite{rubenstein2012kilobot, rubenstein2014programmable}.
We also list one of the NVIDIA robots, NVIDIA JetBot, in the online table, which is widely used for \ac{cav} research and education. 
In the following sections, we discuss the characteristics derived from the sense-plan-act paradigm. 

\section{Characteristics of Testbeds}

In this section, we present and explore a subset of the \characteristicsCount derived characteristics, aiming to offer an overview and discuss some of the key characteristics. First, we will introduce general characteristics that cannot be specifically assigned to the sense, plan, or act domains. Subsequently, we will delve into the characteristics that are unique to each of these domains, providing a more detailed examination of their specific attributes.
For an extensive list containing all \characteristicsCount characteristics, please refer to our online table \cite{webpage}. Note that the characteristics are based solely on objective information and are not intended to rank testbeds.

\subsection{General Characteristics}
The primary design objective for most testbeds is to support users in testing algorithms within a small-scale, controlled environment. Despite the diversity in the specific purposes of individual testbeds, commonalities exist among them. This section aims to outline the general characteristics by highlighting these shared features.

\subsubsection{\textbf{Focus}}
The focus characteristic encapsulates a short, high-level description of a testbed. Typically included in the introduction to the testbed, it describes the specific use cases that the testbed is designed to address. This description serves as a foundational overview, providing a quick insight into the primary objectives of the testbed. In the context of our exploration, we leverage this characteristic to explain the use cases investigated, the methodologies employed, and the insights garnered from the experiments conducted within the testbed. This enables an understanding of testbeds' overarching purpose. Furthermore, testbeds can concentrate on specific problems within an area of focus. For example, within the focus on multi-agent coordination, specific attention could be given to traffic management, communication, and human-autonomy interactions.

\subsubsection{\textbf{Software}}
The diversity in software languages, architectures, frameworks, and computation models across different testbeds reflects the versatility of the testbeds.

Different testbeds may utilize various software components depending on factors such as the targeted application domain, the availability of libraries and tools, and the preferences of the researchers involved. For instance, some testbeds may use Python, a widely adopted language known for its simplicity and extensive scientific computing libraries. Others may use languages such as C++ for performance optimization or MATLAB for domain-specific functionalities.

Another dimension of the software characteristic is the selection of a software architecture. The architecture determines how the software components are organized and communicate with each other. Examples include service-oriented architectures, client-server architectures, peer-to-peer architectures and event-driven architectures. The choice of software architecture impacts factors like scalability, fault-tolerance, modularity, and interoperability within the testbed.

Frameworks also contribute to the software characteristic of testbeds. Frameworks provide reusable software components, libraries, and tools that simplify the implementation of software. Testbeds can leverage existing frameworks tailored for specific purposes, for example, simulation frameworks, Internet of Things (IoT) frameworks, or machine learning frameworks. 

Computation models represent another vital aspect of the software characteristic. Some testbeds employ deterministic computation models such as Finite State Machines for software design, which involve predictable and reproducible outcomes based on defined input conditions. While deterministic models are particularly suitable for experiments that require precise and reliable results, probabilistic computation models introduce randomness and uncertainty into the experiment outcomes, allowing researchers to explore scenarios where stochastic factors play a significant role. Hybrid models combine both deterministic and probabilistic approaches, providing a balance between predictability and stochasticity in the experimental outcomes.

\subsubsection{\textbf{Documentation}}
The documentation characteristic of a testbed refers to the availability and quality of accompanying documentation that describes the testbed's features, functionalities, usage guidelines, and other relevant information. It plays a critical role in the selection of a testbed, as it significantly impacts the user's ability to understand, utilize, and effectively conduct experiments within the given environment.

\subsubsection{\textbf{Accessibility}}
The accessibility of a testbed refers to the ease with which interested parties can access and utilize the facility for conducting experiments. This subsection explores different strategies employed to make testbeds accessible.

Some testbeds provide open-source blueprints that document their design, construction processes, component specifications, and operational procedures in detail, enabling reproduction of their setups. Additionally, some testbeds offer a ready-to-use setup for purchase, which includes all necessary hardware and software, allowing users to quickly begin experiments without the need for extensive technical expertise or complex construction. Such testbeds streamline the experimentation process, particularly for researchers who may not have the necessary expertise or resources to build their own testbed from scratch. Therefore, both open-source sharing and ready-to-use options significantly broaden the testbed's accessibility to a diverse range of researchers.

To accommodate various user needs, testbeds may offer walk-in and remote-access options. Walk-in access allows users to physically engage with the testbed, fostering direct interaction and hands-on experiments.
In contrast, remote-access capabilities enable users to control and monitor experiments from any location via web-based interfaces or specialized software, making the testbed accessible to researchers who are geographically distant or face mobility constraints. In addition, remote access represents a cost-effective method for utilizing testbed resources.
Therefore, both walk-in and remote-access options enhance the accessibility of testbeds, ensuring the efficient utilization of testbed resources and expanding the potential user base.

\subsubsection{\textbf{Scenario}}
A scenario refers to a specific set of conditions, events, and interactions that are implemented within the testbed. Scenarios often involve the deployment of different maps and the presence of diverse actors (e.g., pedestrians, obstacles, etc.), each contributing to the complexity and realism of the experimental setup. The choice of scenarios enables researchers to address specific research questions. Therefore, the selection of a scenario directly correlates with the researchers' focus.

In addition to the above-mentioned characteristics, there are other detailed general characteristics of each testbed in our online table \cite{webpage}, such as the main references related to the testbed, the acquisition cost (if commercially available), and the number of search results in Google Scholar using specific keywords that include the testbed name.

\subsection{Sense-Driven Characteristics}
In the sense-plan-act paradigm, the sense domain primarily focuses on environmental sensing and collection of relevant data via sensors. The key characteristics defining this domain are outlined as follows.

\subsubsection{\textbf{Sensors}}
Sensors form the foundation of perceiving the environment in a testbed. Different sensors, such as cameras, LiDAR, radar, and ultrasonic sensors, provide distinct information about the surrounding objects and their properties. The availability and quality of sensors in a testbed greatly affect the richness and accuracy of the perceptual data collected. Researchers can evaluate the sensor suite and its capabilities to determine whether the testbed adequately supports their sensing needs.

\subsubsection{\textbf{Positioning System}}
Many applications benefit from having accurate and reliable positioning information. A testbed could include an accurate positioning system such as an indoor motion capture system, which is capable of precisely tracking the positions and orientations of agents. 

\subsubsection{\textbf{Accuracy}}
The testbed's sensors supply perceptual information. The accuracy of this information is an aspect to consider. It relates to how closely the sensed data represents the actual state of the environment. Researchers should consider the accuracy of individual sensors and the overall sensing system to assess whether the testbed meets their specific accuracy requirements.

\subsubsection{\textbf{Traffic Management Infrastructure}}
Traffic management infrastructure includes elements such as road networks, traffic signs, traffic lights, and other components that simulate real-world traffic infrastructure. A testbed with a comprehensive traffic management infrastructure enables researchers to study and evaluate the performance of their algorithms and systems in realistic traffic conditions.

\subsubsection{\textbf{Surroundings}}
The environmental surroundings of a testbed contribute to its suitability for experiments. The characteristics of the surroundings, such as urban, suburban, or rural settings, can influence the complexity of the testbed. For instance, an urban testbed might offer challenges related to high-density traffic, while a suburban testbed could have different obstacles and road conditions. 

Other sense-driven characteristics in addition to the ones described above are available in our online table \cite{webpage}. We also expand on several characteristics to provide in-depth information. For instance, we distinguish between on-agent and global sensors, offering a more comprehensive understanding of the sensor suite employed within a testbed.

The detailed exploration of sense-driven characteristics provides a solid foundation for assessing a testbed's capabilities in environmental sensing. These sensory inputs are used within the plan domain to enable effective planning.

\subsection{Plan-Driven Characteristics}
The plan domain plays a crucial role in enabling \domainName{} to generate appropriate actions based on the information gathered from the environment. 
This section focuses on deriving characteristics related to the plan domain that can be used to evaluate and compare testbeds. 

\subsubsection{\textbf{Testbed Architecture}}
The testbed architecture refers to the overall design and structure of the experimental setup. It encompasses the physical layout, the presence of specific infrastructure elements, and the integration of hardware and software components. When evaluating a testbed for the plan domain, researchers should consider whether the architecture supports the execution and evaluation of planning algorithms and strategies effectively. For reproducible experiments, a deterministic timing of the testbed components is required, i.e., the clocks of all computing and sensing devices have to be synchronized. 

\subsubsection{\textbf{Distributed Computation}}
In many real-world scenarios, \domainName{} operate in distributed environments, where the processes are spread out across multiple nodes. Assessing the capability of a testbed to handle distributed computation is essential, as it determines whether the testbed can efficiently manage coordination and communication among agents during the planning phase.

\subsubsection{\textbf{Computation Unit}}
Computation units are tasked with processing sensor data and executing complex algorithms, making them integral components in testbeds. They are essential due to their role in handling real-time planning processes, ensuring operational efficiency and reliability in dynamic environments. Additionally, the adaptability of computation units to new software and technologies is crucial for future-proofing testbeds. Thus, computation units should be carefully considered when selecting or developing testbeds.

\subsubsection{\textbf{Computation Schemes}}
The computation scheme refers to the methodology or algorithms employed for planning within the testbed. Distributed planning algorithms follow different computation schemes, e.g., sequential, parallel, and hybrid computations. Hybrid computations involve a combination of sequential and parallel processing methods to optimize planning within a testbed.

\subsubsection{\textbf{Human-Robot Interaction}}
In many real-world scenarios, \domainName{} coexist and interact with humans in shared environments. Evaluating the testbed's characteristics related to human-robot interaction is essential for assessing its suitability for scenarios involving mixed traffic. Characteristics such as the availability of realistic human models, capabilities for simulating human behaviors, or an interface for humans to interact with the testbed are important considerations.
  
\subsubsection{\textbf{Different Kinds of Agents}}
Assessing the ability of a testbed to accommodate heterogeneous agents is vital when studying \domainName. Different types of agents may have distinct sensing and acting capabilities, leading to diverse planning requirements. A testbed should support the integration of various agent models, allowing researchers to evaluate planning algorithms for scenarios involving agents with different capabilities, such as cars and trucks. 

\subsubsection{\textbf{Agent Count}}
The number of agents in a testbed is an important characteristic, particularly in scenarios involving transportation or traffic management. Evaluating the agent count characteristic involves understanding the testbed's capacity to manage multiple agents concurrently, the scalability of planning algorithms with increasing agent count, and the  impact on the planning system's performance as the number of agents increases.

\subsection{Act-Driven Characteristics}
This subsection aggregates information related to translating a plan into action, which in \domainName{} applications often entails trajectory following by agents.
In a hierarchical architecture such as the sense-plan-act paradigm, planning is a separate layer from trajectory following. An agent model links the two layers by representing constraints on the agent dynamics.

\subsubsection{\textbf{Dynamics}}
Modeling the system dynamics is a fundamental step in classical control design. When  planning respects the constraints on the agent dynamics, more accurate trajectory following is possible. If the agent model significantly deviates from the actual agent dynamics, a theoretically collision-free trajectory could lead to a collision when executed. Hence, it is crucial to test the trajectory planning model through experiments with real agents.
Many \domainName{} agents can be categorized as differential-drive or Ackermann-steering, depending on their steering geometry.
Models for differential-drive agents include the point-mass model with double-integrator dynamics and the unicycle model, which separates control inputs for translational and rotational velocity \cite{schwab_experimental_2020, pickem2017robotarium}.
Models for Ackermann-steering agents include the point-mass model with double-integrator dynamics, the kinematic bicycle model, and the kinetic bicycle model. An overview of agent models is given in \cite{althoff2017commonroad}.
Depending on the model choice, different agent parameters must be known.
Generally speaking, the limits on an agent's speed and acceleration affect the constraints on its longitudinal dynamics, and
the limits on its steering angle and steering rate affect the constraints on its lateral dynamics.

\subsubsection{\textbf{Geometry}}
The models for agent dynamics given above describe the motion of an agent's center of gravity.
The occupied area of an agent is distributed around this center of gravity. In order to ensure collision-free plans, we need to guarantee that no occupied areas intersect. The agent's geometry is needed to compute its occupied area.
\subsubsection{\textbf{Battery}}
Batteries are important components to supply energy for agents, and they contribute significantly to the total agent weight. A suitable battery selection is essential to maintain a balance between battery runtime and agent mobility, particularly for compact or agile agents. However, the selection should be aligned with the particular uses of the testbed. For example, while testbeds for education may demand a relatively long runtime with one full charge of agents if many educational activities take place on the testbed, testbeds for racing should prioritize a low ratio of battery weight to agent weight over a long battery runtime. Therefore, our online table includes information about the batteries used on each agent, including the battery runtime and the battery-to-agent weight ratio. 

\section{Overview of Testbeds}
\begin{figure*}[t] 
    \centering
    \begin{minipage}{0.48\textwidth}
        \centering
        \includegraphics[width=\columnwidth]{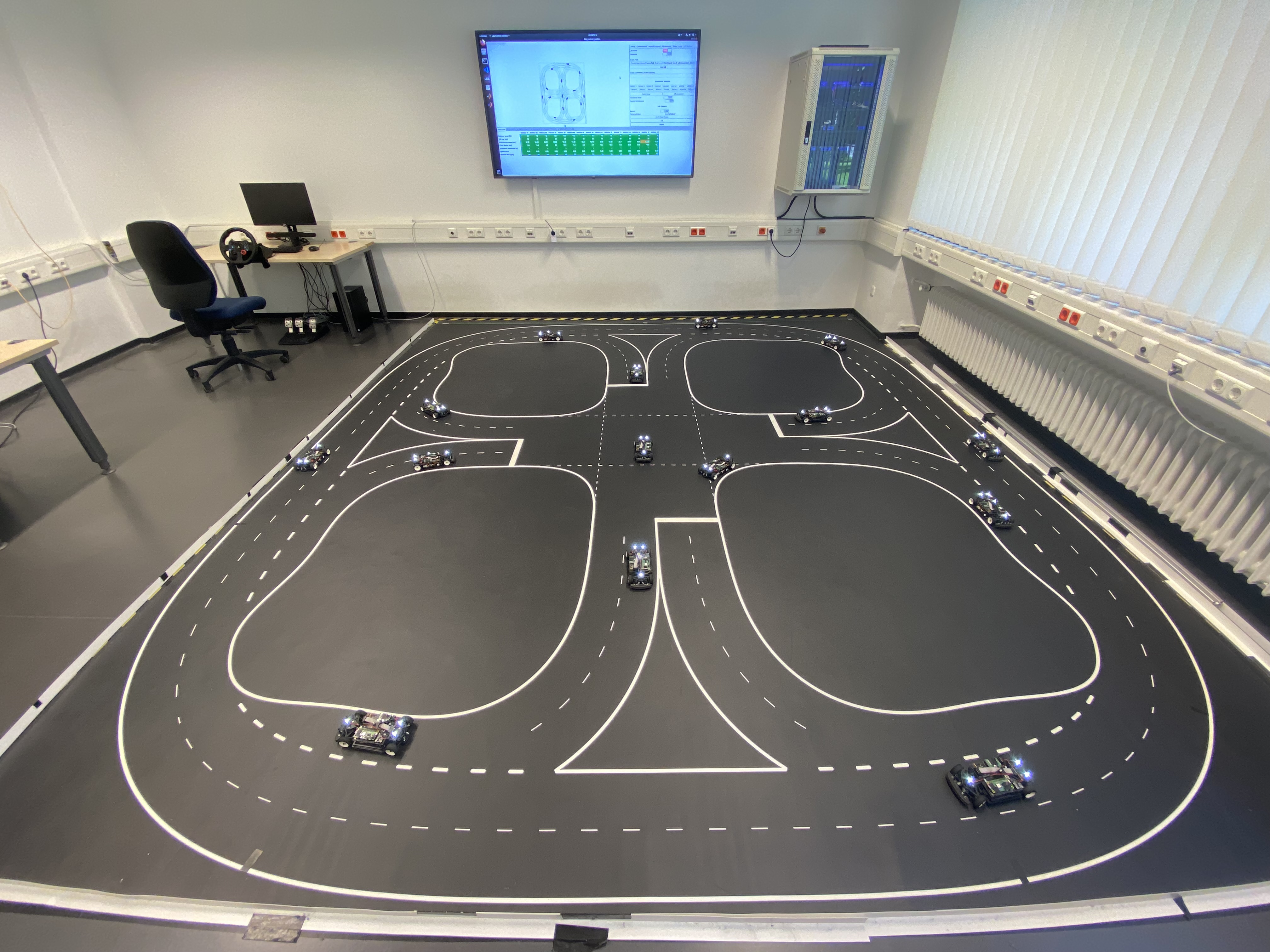}
        \caption{\mbox{Cyber-Physical Mobility Lab}, originally developed at RWTH Aachen University, now moved to University of the Bundeswehr Munich \cite{kloock_cyber-physical_2021}.}
        \label{fig:cpmLab}
    \end{minipage}%
    \hfill
    \begin{minipage}{0.48\textwidth}
        \centering
        \includegraphics[width=\columnwidth]{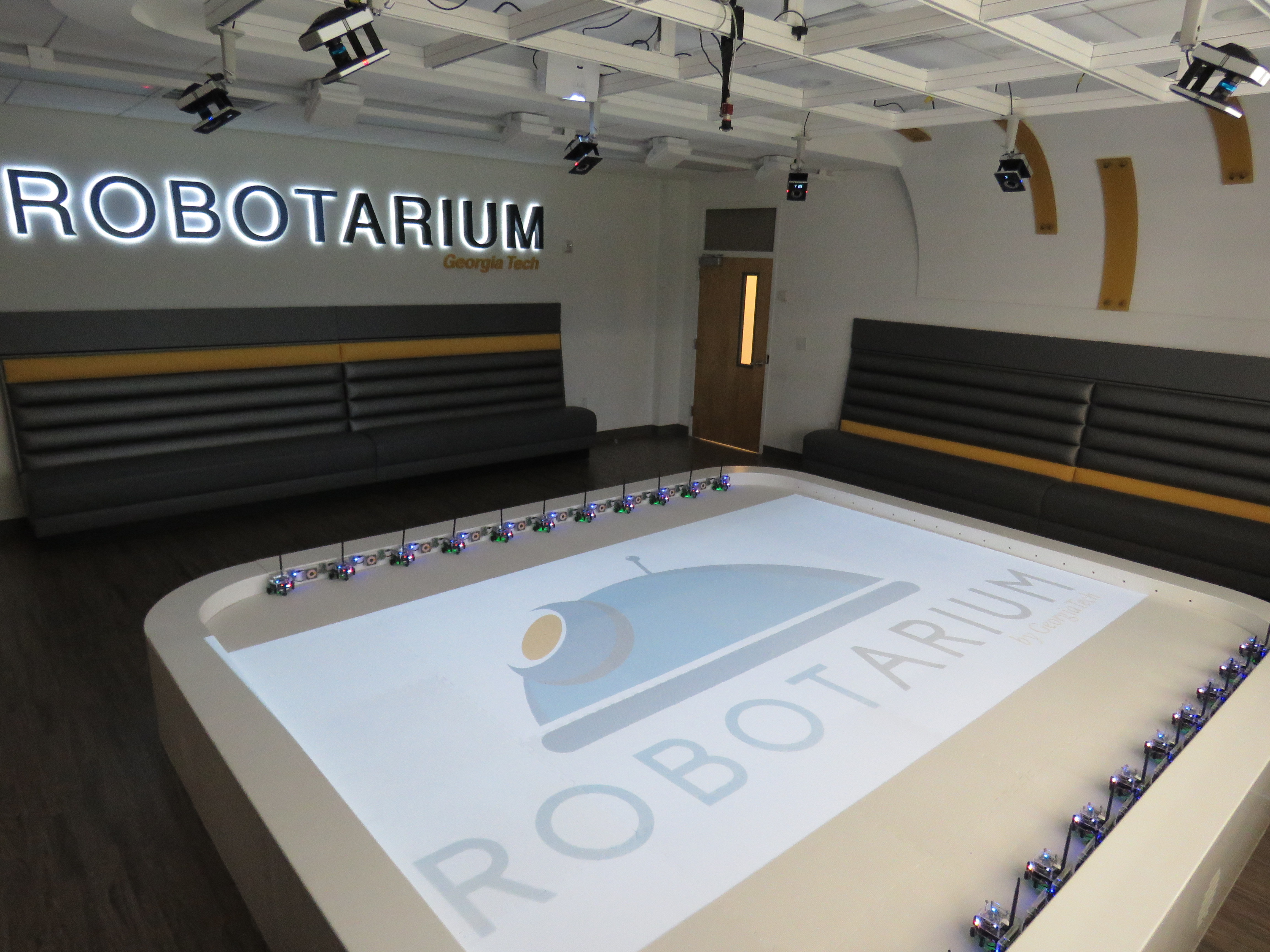}
        \caption{Robotarium testbed at Georgia Institute of Technology \cite{pickem2017robotarium}.}
        \label{fig:ROBO}
    \end{minipage}
    \begin{minipage}{0.48\textwidth}
        \centering
        \includegraphics[width=\columnwidth]{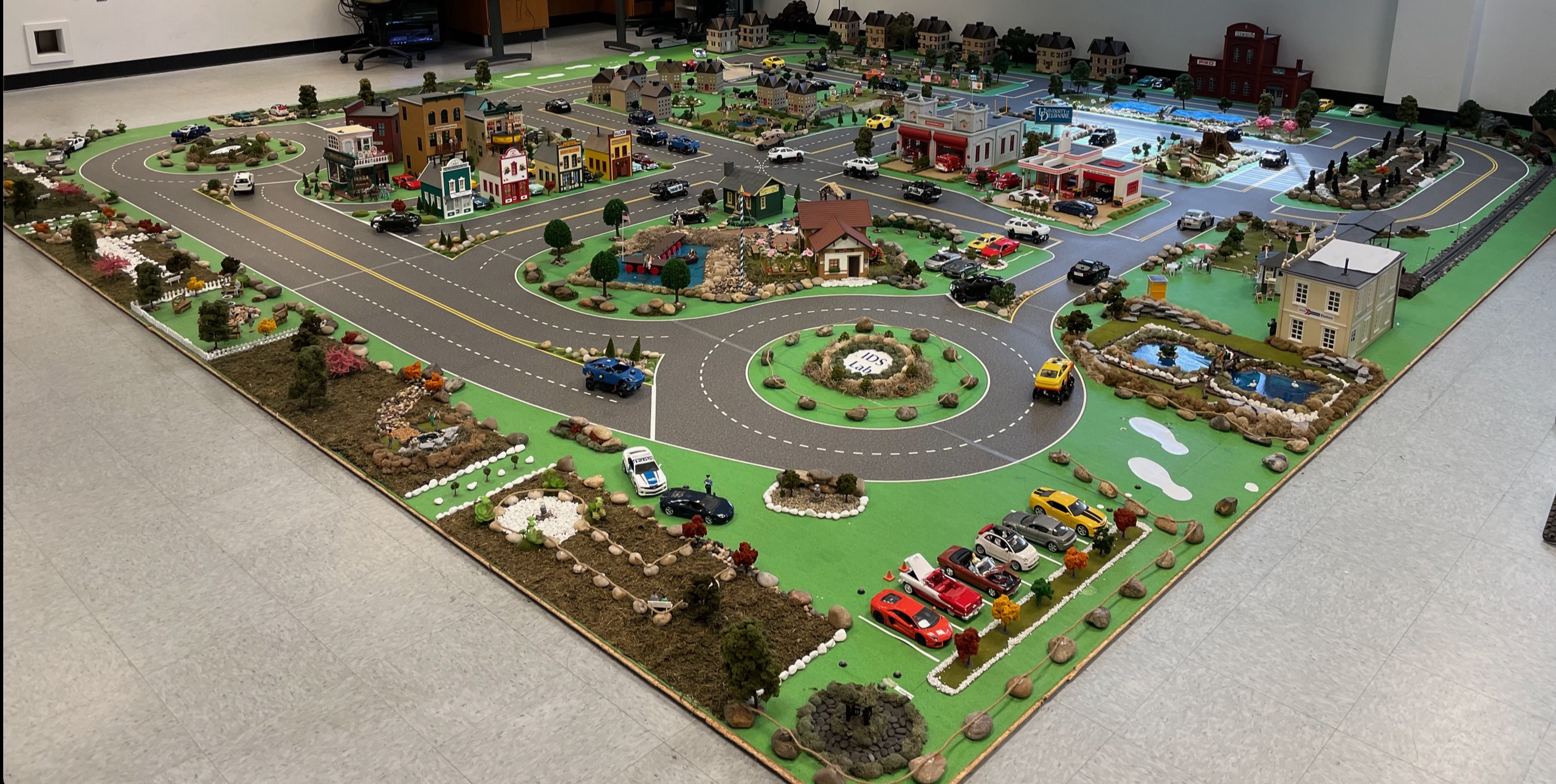}
        \caption{Information and Decision Science Lab Scaled Smart City (IDS3C) at Cornell University \cite{stager_scaled_2018}.}
        \label{fig:IDSLab}
    \end{minipage}%
    \hfill
    \begin{minipage}{0.48\textwidth}
        \centering
        \includegraphics[width=0.90\columnwidth]{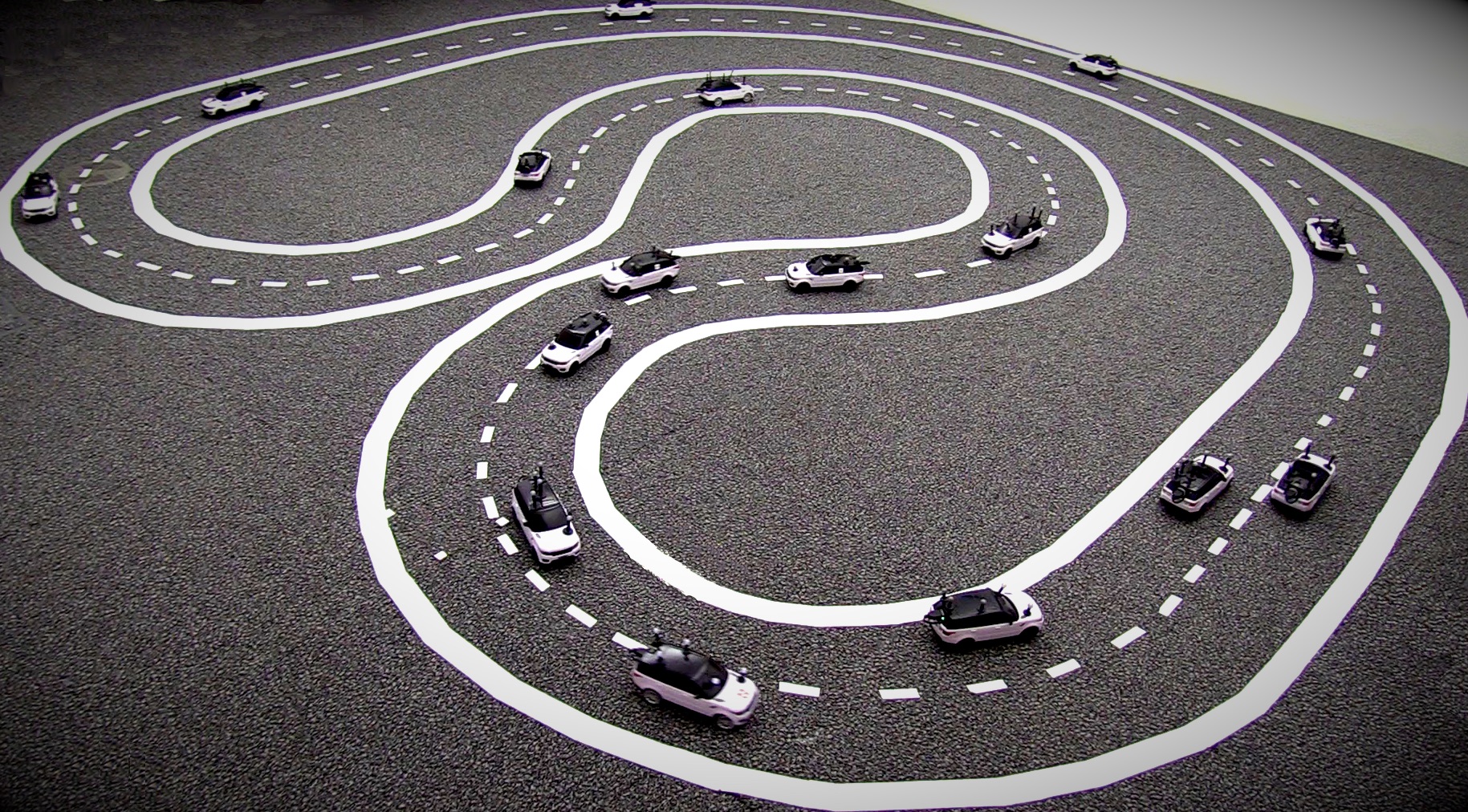}
        \caption{Cambridge Minicar testbed at the Prorok Lab at Cambridge University \cite{hyldmar_fleet_2019}.}
        \label{fig:Prprok}
    \end{minipage}
\end{figure*}
\begin{figure*}[t]
    \centering
    \begin{minipage}{0.48\textwidth}
        \centering
        \includegraphics[width=0.8\columnwidth]{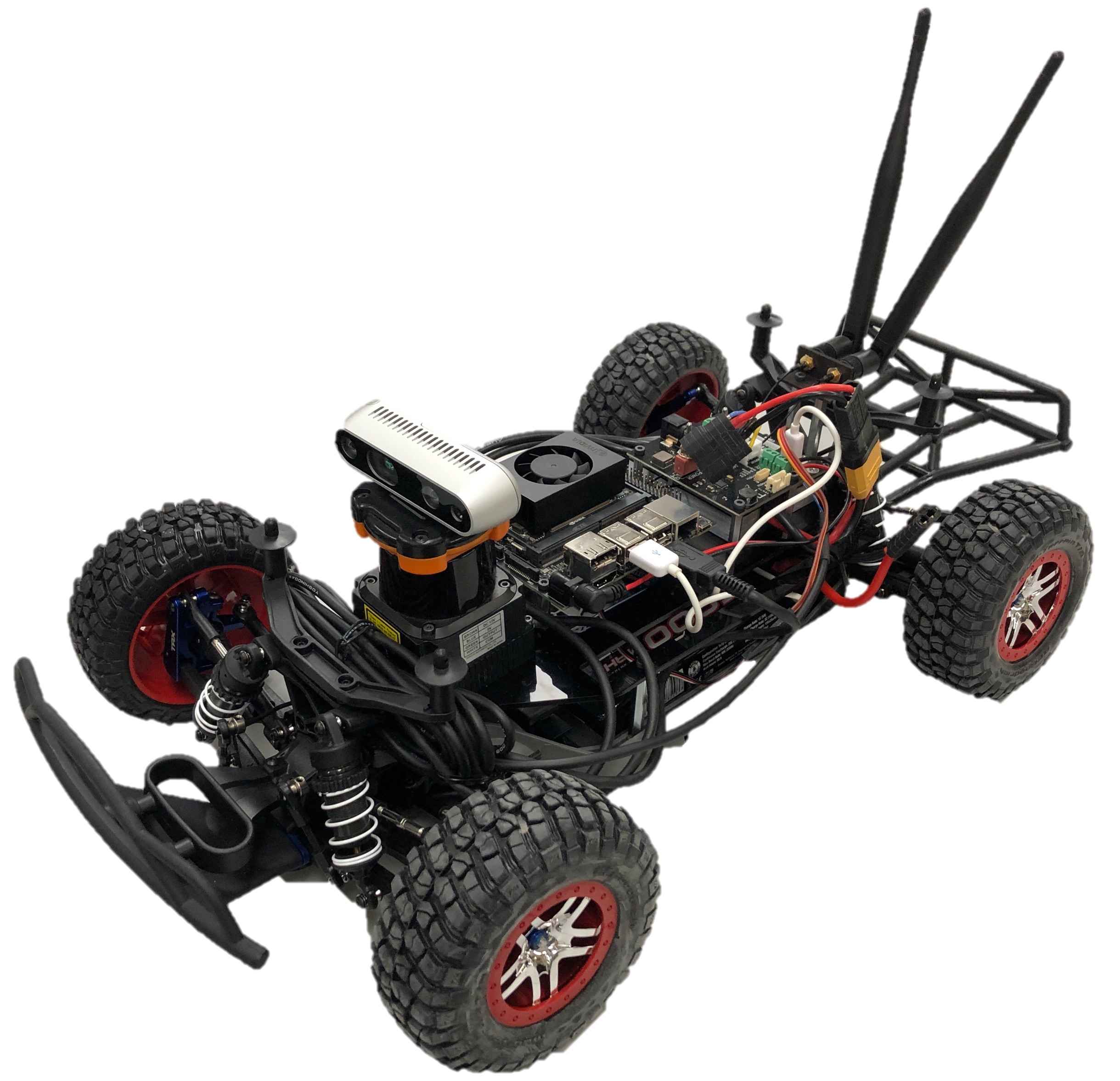}
        \caption{An exemplar F1TENTH agent built at the University of Pennsylvania \cite{okelly_f1tenth_2020}.}
        \label{fig:F1TENTH}
    \end{minipage}%
    \hfill
    \begin{minipage}{0.48\textwidth}
        \centering
        \includegraphics[width=\columnwidth]{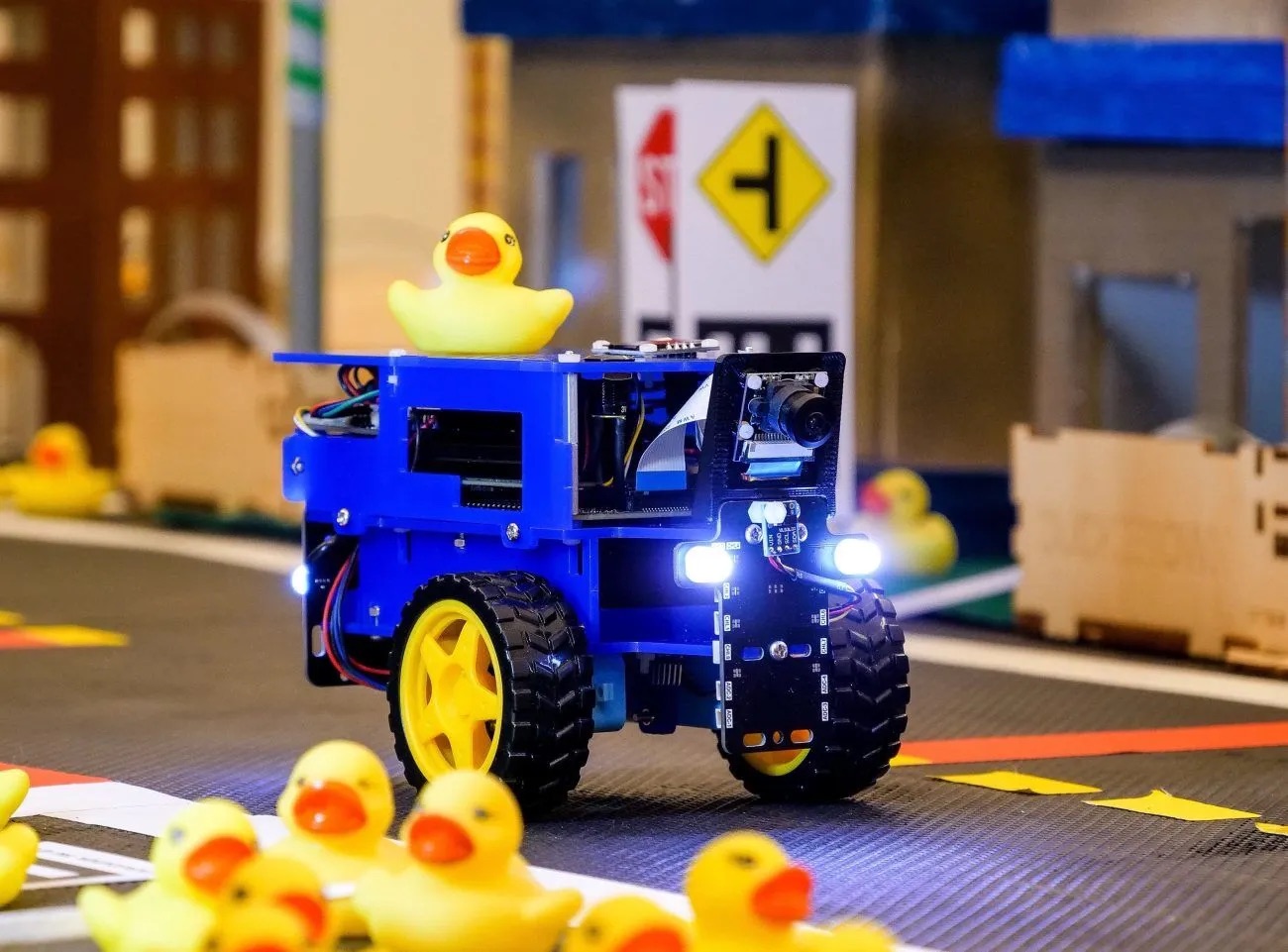}
        \caption{Duckietown testbed, originally developed at Massachusetts Institute of Technology \cite{paull_duckietown_2017}.}
        \label{fig:Duckie}
    \end{minipage}
    \vspace{1ex} 
    \begin{minipage}{0.48\textwidth}
        \centering
        \includegraphics[width=\columnwidth]{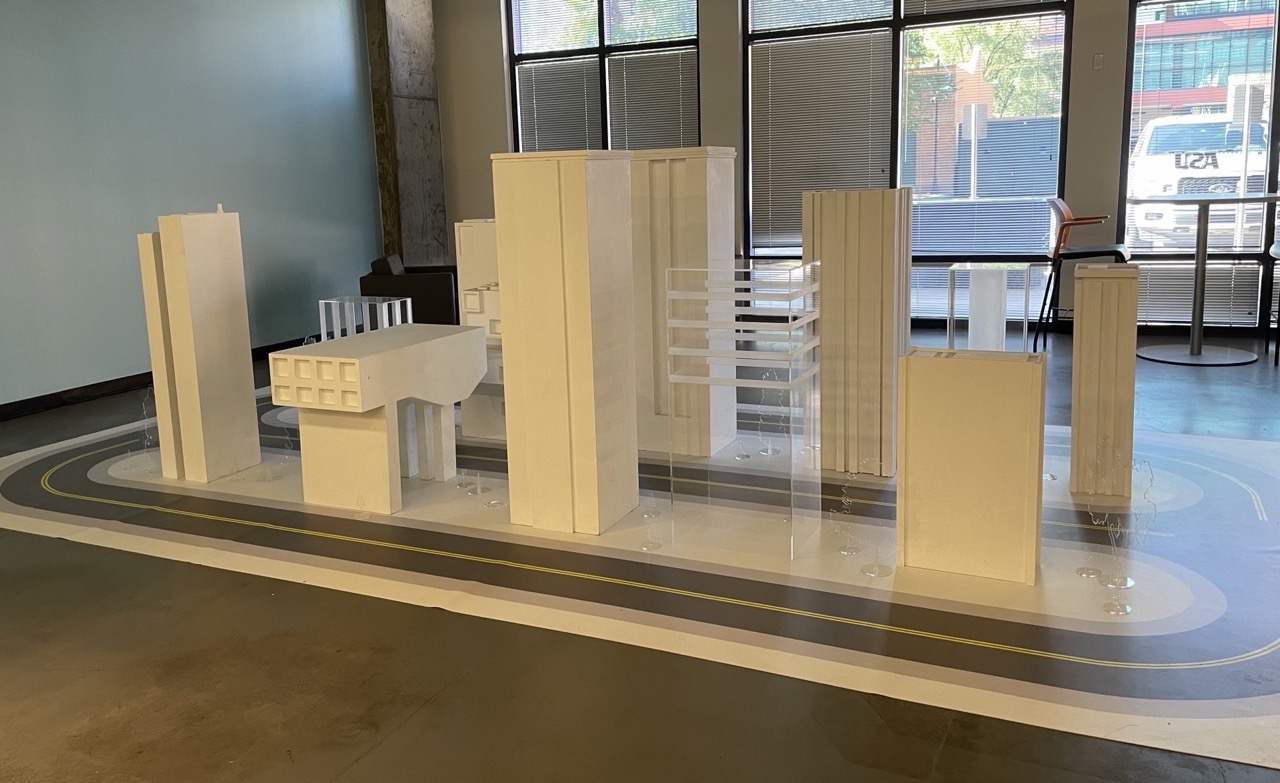}
        \caption{CHARTOPOLIS testbed at Arizona State University \cite{ulhas_chartopolis_2022}.}
        \label{fig:CHARTOPOLIS}
    \end{minipage}%
    \hfill
    \begin{minipage}{0.48\textwidth}
        \centering
        \includegraphics[width=0.88\columnwidth]{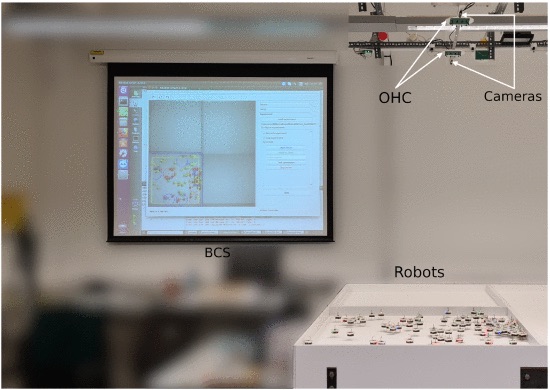}
        \caption{Augmented Reality for Kilobots (ARK) testbed at University of Sheffield \cite{reina2017ARK}.}
        \label{fig:ARK}
    \end{minipage}
\end{figure*}
\begin{figure}
    \centering
    \includegraphics[width=\linewidth]{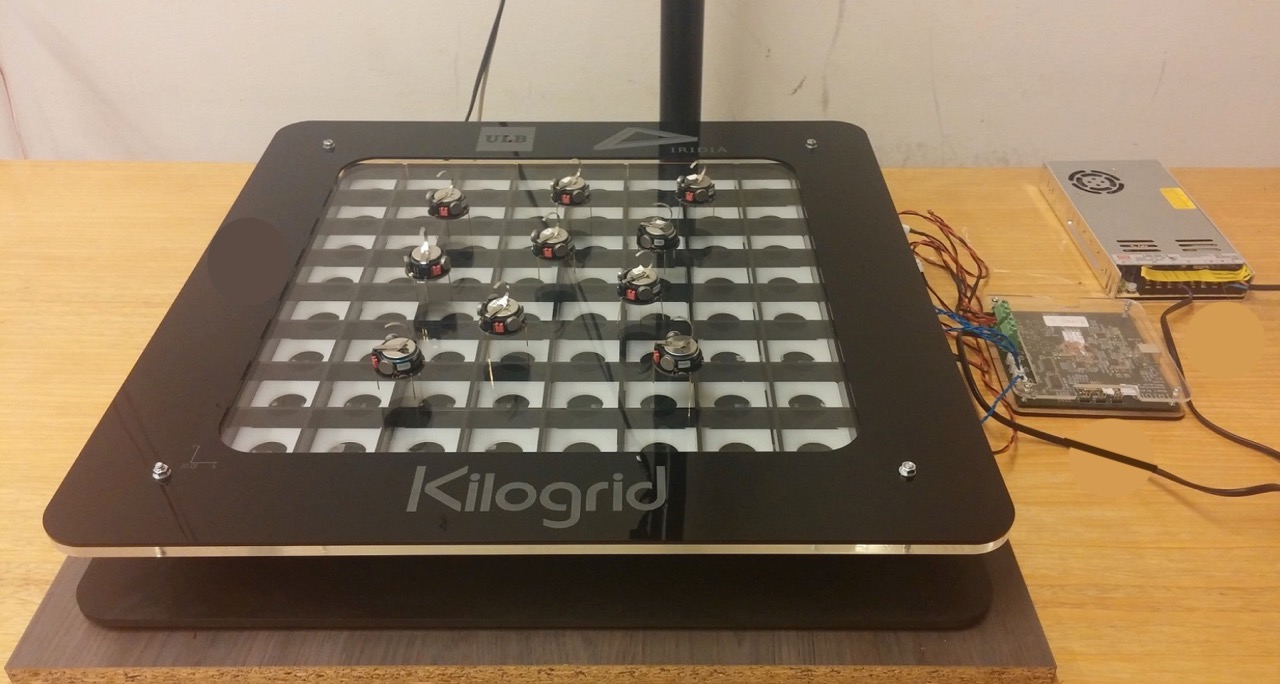}
    \caption{Kilogrid testbed at Free University of Brussels \cite{antoun2016kilogrid}.}
    \label{fig:kilogrid}
\end{figure}
Through this paper, we aim to provide researchers with an understanding of the current landscape of \domainName{} testbeds. By examining the sense-plan-act paradigm characteristics, we gain deeper insights into the capabilities, strengths, and potential areas of advancement of each testbed. 
As discussed in Section~\ref{sec:contributions}, 
we select \mainTestbeds out of \TestbedCount testbeds and will examine them in depth in this section.
In addition, we provide Table \ref{tab:testbeds_research_areas} to help readers quickly identify the main research areas of all  \TestbedCount testbeds. We categorize them into five areas: planning and control, computer vision, collective behavior, autonomous racing, and human-robot interaction.

\renewcommand{\arraystretch}{1.5}  

\begin{table}[t!]
    \centering
    \caption{Main research areas of the \TestbedCount testbeds.}\label{tab:testbeds_research_areas}
    \begin{tabular}{l c c c c c}
        \toprule
        & \rotatebox{90}{\parbox{1.7cm}{Planning and\\ Control}} 
        & \rotatebox{90}{\parbox{1.7cm}{Computer\\ Vision}} 
        & \rotatebox{90}{\parbox{1.7cm}{Collective\\ Behavior}} 
        & \rotatebox{90}{\parbox{1.7cm}{Autonomous\\ Racing}} 
        & \rotatebox{90}{\parbox{1.7cm}{Human-Robot\\ Interaction}} \\
        \midrule
        CPM Lab \cite{kloock_cyber-physical_2021} & \checkmark &  &  &  & \checkmark \\
        F1TENTH \cite{okelly_f1tenth_2020} & \checkmark & \checkmark &  & \checkmark &  \\
        Robotarium \cite{pickem2017robotarium} & \checkmark &  & \checkmark &  &  \\
        IDS3C \cite{stager_scaled_2018} & \checkmark &  &  &  & \checkmark \\
        Cambridge Minicar \cite{hyldmar_fleet_2019} & \checkmark &  &  &  & \checkmark \\
        Duckietown \cite{paull_duckietown_2017} & \checkmark & \checkmark &  &  &  \\
        CHARTOPOLIS \cite{ulhas_chartopolis_2022} &  \checkmark & \checkmark &  &  & \checkmark \\
        ARK \cite{reina2017ARK} & \checkmark &  & \checkmark &  &  \\
        Kilogrid \cite{antoun2016kilogrid} & \checkmark &  & \checkmark &  &  \\
        Cambridge RoboMaster \cite{blumenkamp2024cambridge} & \checkmark & \checkmark &  &  &  \\
        Go-CHART \cite{kannapiran_go-chart_2020} &  & \checkmark &  &  & \checkmark \\
        CRS \cite{carron_chronos_2023} & \checkmark &  &  & \checkmark &  \\
        Pheeno \cite{wilson_pheeno_2016} &  & \checkmark & \checkmark &  &  \\
        MiniCity \cite{buckman_evaluating_2022} & \checkmark & \checkmark &  &  &  \\
        SAMS \cite{schwab_experimental_2020} & \checkmark &  &  &  &  \\
        CPS Lab \cite{graham_abstractions_2009} & \checkmark &  &  &  &  \\
        MCCT \cite{dong_mixed_2023} & \checkmark &  &  &  &  \\
        UPBOT \cite{crenshaw_tanya_l_and_beyer_steven_upbot_2010} & \checkmark &  &  &  &  \\
        ORCA \cite{liniger_optimization-based_2015} & \checkmark &  &  & \checkmark &  \\
        Miniature Autonomy \cite{tiedemann_miniature_2022} & \checkmark & \checkmark &  &  &  \\
        ICAT \cite{tian2024icat} & \checkmark & \checkmark &  &  &  \\
        ETLMT \cite{michael2008experimental} & \checkmark &  & \checkmark &  &  \\
        MiniCCAM \cite{wang2024introduction} & \checkmark &  & \checkmark &  & \checkmark  \\
        \bottomrule
    \end{tabular}
\end{table}

In the following, we use the selected testbeds to illustrate how the derived characteristics can be used to describe a testbed effectively. 
The selected testbeds are:
the Cyber-Physical Mobility Lab (\mbox{CPM Lab}) \cite{kloock_cyber-physical_2021}, 
F1TENTH \cite{okelly_f1tenth_2020}, 
the Robotarium \cite{pickem2017robotarium}, 
IDS3C \cite{stager_scaled_2018}, 
the Cambridge Minicar~\cite{hyldmar_fleet_2019}, 
Duckietown \cite{paull_duckietown_2017}, 
\mbox{CHARTOPOLIS} \cite{ulhas_chartopolis_2022}, 
ARK \cite{reina2017ARK}, 
and
Kilogrid \cite{antoun2016kilogrid}.
Figures \ref{fig:cpmLab} to \ref{fig:kilogrid} show these testbeds, respectively.
To ensure transparency, we provide the affiliations of all authors to the selected testbeds at the application part of this paper.

\subsection{General Information}
This section discusses the specific application of the general characteristics derived in the previous section, i.e., focus, software, documentation, accessibility, and scenario, to characterize the \mainTestbeds testbeds.

\subsubsection{\textbf{Focus}}
The Cambridge Minicar, the IDS3C, \mbox{CHARTOPOLIS}, and the \mbox{CPM Lab} are 
designed for multi-agent planning and control of \acp{cav}. In addition, they can also be used to investigate interactions between \acp{cav} and human-driven vehicles. For example, two driving modes are being developed for \mbox{CHARTOPOLIS}, as they were for its smaller predecessor described in \cite{kannapiran_go-chart_2020}: one in which a vehicle navigates autonomously, and one in which it is remotely driven by a human participant. In contrast, testbeds such as Duckietown and F1TENTH primarily focus on sensing. Duckietown implements a miniature-city-scale environment for exploring camera-based localization, navigation, and coordination, while F1TENTH operates on a larger scale (1/10) and employs a LiDAR sensor.

While the aforementioned testbeds target \acp{cav}, the Robotarium, ARK, and Kilogrid are designed for \acp{rs} and facilitate the study of distributed algorithms in swarm robotics, a field focused on the coordinated behavior and interactions of multiple autonomous agents operating as a collective system.
Specifically, the Robotarium provides a remotely accessible swarm robotics research testbed. It gives users the flexibility to test a variety of multi-robot algorithms and addresses the concern of safety with formal methods to avoid damage to the agents.
Kilobot \cite{rubenstein2012kilobot, rubenstein2014programmable} is a small robot designed to study the collective behaviors of large-scale autonomous swarms. It is made with low-cost parts and mostly assembled by an automated process, enabling mass production. While cost-effective, Kilobots are only equipped with a minimal set of sensors and actuators, which limits the range of tasks they can perform.
Two testbeds have been developed that enhance the Kilobots' capabilities with virtual sensors and actuators using augmented reality.
One of these testbeds, the Kilogrid \cite{antoun2016kilogrid, valentini2018kilogrid}, is a modular and scalable virtualization environment designed for the study of collective behaviors in swarm robotics.
The other testbed, ARK \cite{reina2017ARK}, also uses augmented reality to enhance Kilobots with virtual sensors and actuators; it is more cost-effective and can be assembled more quickly than the Kilogrid.
Moreover, while the Kilogrid operates on discrete cell modules, ARK tracks robots in continuous space, which favors more precision tasks.
Additionally, ARK has been applied to another type of agent called the e-Puck \cite{mondada2009puck}, which is a widely-used robot designed originally for education in engineering.

Although most of the testbeds are primarily utilized for research purposes, some are also integrated into educational settings and employed for competitions.
In particular, F1TENTH has been utilized in university courses focused on autonomous vehicles \cite{betz2022teaching}, Duckietown offers an online open course on AI and robotics \cite{duckietown-mooc}, and the CPM Academy \cite{mokhtarian_cpm_2023, mokhtarian_remote_2022, scheffe2020networked} allows users to learn about and address specific problems in intelligent transportation systems.
Both Kilobots and e-Pucks have also been adopted into university courses to teach swarm intelligence.
In terms of competitions, F1TENTH is used for autonomous racing competitions that challenge participants to avoid crashes and minimize lap times \cite{f1tenth-race}. Duckietown hosts the AI Driving Olympics, a semi-annual competition that focuses on machine learning and artificial intelligence \cite{ai-driving-olympics}. The \mbox{CPM Lab} introduced the annual CPM Olympics \cite{mokhtarian2022cpm}, which enables users to remotely access challenging real-world \ac{cav} scenarios deployed on the testbed. 

\subsubsection{\textbf{Software}}
Some testbeds are built upon common open-source software architectures. For instance, IDS3C and F1TENTH employ the Robot Operating System (ROS) and the \mbox{CPM Lab} utilizes the Data Distribution Service (DDS). Duckietown has developed a containerized architecture, in which the system is divided into smaller, isolated units called containers. Each container leverages ROS and is responsible for a specific part of the main functionalities.
The Robotarium follows a comparable approach, where the architecture consists of three main groups: simulation-based components, testbed interface components, and coordinating server applications. Similarly, \mbox{CHARTOPOLIS} uses a modified Donkey Car \cite{autorope/donkeycar} library or the ROS library based on the application requirements. 
The Kilogrid software architecture is centered on an application called KiloGUI, which enables users to configure, program, and manage Kilogrid modules and the agents. Key functionalities include uploading and running programs, logging data, and setting up the operational environment through the GUI. Additionally, the software includes a dispatcher component that handles communication between the modules and KiloGUI.
The software of the ARK testbed is centered around the Base Control Software (BCS), which coordinates agents in augmented reality setups. It processes images from multiple cameras to construct a unified arena view, tracks the positions and states of individual agents, and manages communication via infrared signals. The BCS employs multi-threading and utilizes GPU processing to maintain operational efficiency and manage real-time tasks. The software also features a user interface for setting up and monitoring experiments. Additionally, it enables the creation of virtual environments that expand the experimental capabilities by simulating different sensory inputs.
Concerning the supported programming languages, the selected testbeds utilize some widely-used options, such as C/C++, Python, and/or MATLAB.

\subsubsection{\textbf{Documentation}}
The majority of the testbeds provide users with comprehensive manuals and access to source code. F1TENTH and the \mbox{CPM Lab} offer complete open-source code that facilitates the reconstruction and comprehension of the testbed's architecture. Conversely, the Robotarium and the Cambridge Minicar offer code that is specifically designed for simulating simple examples. Furthermore, F1TENTH and the \mbox{CPM Lab} furnish a construction manual detailing the process of building vehicles and environments. Duckietown offers an operation manual describing its appearance specifications and assembly instructions, as well as code documentation, 
benchmarks of a set of its competitions called AI Driving Olympics, and 
instructions for developing new educational exercises, among other things. 
The Kilogrid provides documentation containing the basic information required to get started with the Kilogrid system. ARK provides an open-source tool with full documentation to manage and edit all components of the system.
   
\subsubsection{\textbf{Accessibility}}
Each testbed offers varying degrees of accessibility. 
The Robotarium and the \mbox{CPM Lab} 
enable users to remotely interact with the testbeds through the internet. In addition, the \mbox{CPM Lab} has seamlessly integrated its entire interface, including a simulation environment, within a web application. Although the Robotarium necessitates a local installation for development purposes, its interface can still be accessed through a web application.

To facilitate the utilization of some testbeds, simulators have been developed for the IDS3C, Duckietown, Robotarium, F1TENTH, and \mbox{CPM Lab}. The simulators provide users with the opportunity to test their algorithms without the inherent risks associated with hardware experimentation. The simulators have varying levels of complexity. For example, the \mbox{CPM Lab} includes digital twin representations of the testbed, which are virtual replicas that mimic the behavior and characteristics of the physical system they represent. 
Moreover, there is a substantial variation in the technologies employed, contributing to the diverse complexity of these simulators. IDS3C has introduced a Unity-based simulator known as IDS 3D City \cite{Ray2021DigitalCity}, which allows users to rapidly iterate their control algorithms and experiments before deploying them to IDS3C. The Robotarium provides a simulator compatible with both MATLAB and Python interfaces. \mbox{CHARTOPOLIS} employs a map in the CARLA \cite{Dosovitskiy17} simulation environment to mimic the testbed in simulation. Additionally, Duckietown and F1TENTH have developed simulators built upon the OpenAI Gym framework \cite{duckietown-gym,f1tenth-gym}. Notably, F1TENTH expands its simulation capabilities by providing a simulator integrated within the ROS Gazebo environment \cite{okelly_f1tenth_2020}. These simulators empower users to implement and evaluate their algorithms within meticulously controlled virtual environments.

An alternative approach to enhancing testbed accessibility has been adopted by Duckietown,
aiming to make the setup more affordable and easily replicable. In the case of Duckietown, users have the option to purchase ready-to-use setups, reducing the barriers to entry and facilitating broader adoption.

Although many of the testbeds are not commercially available, the majority of them are open-source, except for IDS3C and the CHARTOPOLIS. Therefore, users can build most of them on their own. We also detail the acquisition costs in our online table \cite{webpage}. For example, the Kilogrid consists of a grid of modules, which can be programmed by users to define complex functionalities. A single module costs approximately \$80, while 200 modules accommodates up to 100 agents \cite{valentini2018kilogrid}. In comparison, ARK costs approximately \$700 as a whole \cite{valentini2018kilogrid}. Note that the acquisition cost of the agents (Kilobots) is the same for both testbeds.

\subsubsection{\textbf{Scenario}}
IDS3C primarily focuses on urban driving scenarios such as intersections, roundabouts, merging roadways, and corridors \cite{chalaki2020experimental,Beaver2020DemonstrationCity}. 
It is also equipped with driver emulation stations (remote vehicle operation), which enables the exploration and study of human driving behaviors and their interaction with \acp{cav}.
The Cambridge Minicar testbed specializes in providing a multi-lane freeway scenario. In contrast to these, F1TENTH, Duckietown, and the Robotarium offer users the ability to customize maps and scenarios according to their preferences. The map of the \mbox{CPM Lab} features an eight-lane intersection and a loop-shaped highway with multiple merge-in and -outs, offering a range of challenging traffic conditions. The CPM Lab also allows users to customize the map to suit their needs, provides real-world scenarios for a competition, and supports another benchmark framework called CommonRoad \cite{althoff2017commonroad}.
The IDS3C and Duckietown enable users to validate traffic management algorithms, as they are equipped with stop and yield signs to simulate realistic traffic conditions.

While the above scenarios require physical hardware, the Kilogrid and ARK testbeds introduce a virtual environment with virtual sensors and actuators for the Kilobot agents, enabling various virtual scenarios such as virtual pheromone trails, gradient sensing, and foraging.

\subsection{Sense-based Information}
Testbeds often vary in their approaches to sensing and localization.
Localization can be accomplished through onboard agent sensors or sensors embedded in the environment. A notable characteristic of most of the investigated testbeds is that they combine both methods. 

\subsubsection{\textbf{Sensors}}
The Duckietown testbed exemplifies agent-centric localization in which the agents rely on an array of onboard sensors, including a Hall effect sensor (odometer), a front-facing camera, a time-of-flight sensor, and an Inertial Measurement Unit (IMU). These sensors collect data on the agent's movement, the surrounding environment, and the agent's position relative to that environment, allowing each agent to estimate its position and independently navigate within the testbed. In addition, the Duckietown ``Autolab'' provides a low-cost external camera-based localization infrastructure \cite{9341677}. 

Similarly, agents in \mbox{CHARTOPOLIS} use onboard cameras for self-localization, and the testbed includes an overhead OptiTrack camera system for enhanced vision-based tracking of agents. However, it is currently not fully integrated, leaving room for future upgrades.
In contrast, the F1TENTH testbed employs a different sensing modality, using a LiDAR sensor mounted on each agent. LiDAR enables the agents to construct a detailed 3D map of their surroundings and precisely determine their location within this map, providing a high level of environmental awareness and localization accuracy.

However, not all the testbeds rely primarily on agent-centric localization. Other testbeds, such as the IDS3C and \mbox{CPM Lab}, demonstrate a globally coordinated approach. While the agents in these testbeds are equipped with sensors such as an IMU and an odometer, these only play a supplementary role in localization. A global positioning system provides the primary means of localization, with further details covered in the subsequent section.

Unlike the physical sensors used in other testbeds, the Kilogrid and ARK testbeds use virtual sensors to create a virtual sensing environment. 
The Kilogrid incorporates a network of interconnected modules that each function as a computing node, providing the Kilobots with the ability to interact with virtual elements that are not present in their physical environment. 
The ARK testbed employs a system of overhead cameras and a modified overhead emitter to create a virtual sensing environment. This system tracks the real-time location and state of each Kilobot and 
communicates location-and-state-based information 
directly to the agents. 
The setups of both testbeds enable the Kilobots to sense and interact with virtual elements such as gradients, reference points, or pheromone trails, which are otherwise intangible. By providing a virtual layer of interaction, they allow researchers to conduct complex experiments that extend beyond the physical capabilities of the Kilobots.

\subsubsection{\textbf{Positioning System}}
\mbox{CPM Lab} is an example case of using a global positioning system for localization. It employs a ceiling-mounted Basler camera that detects LEDs mounted on each agent, where three LEDs indicate the agent's pose and a fourth blinks at a specific frequency to uniquely identify each agent. This LED-based system serves as the primary method for localizing the agents within the \mbox{CPM Lab} environment.

Similar to \mbox{CPM Lab}, the IDS3C, Robotarium, and Cambridge Minicar testbeds use high-precision camera systems to track agent poses. While the IDS3C and the Robotarium use a Vicon camera system, the Cambridge Minicar testbed employs an OptiTrack camera system.
These cameras capture the agents' locations at high refresh rates, enabling accurate real-time localization.
The ARK testbed also employs a centralized system using four overhead cameras to track the real-time location and state of each Kilobot.

In addition to the primary global positioning system, \mbox{CPM Lab} integrates a secondary system based on a sensitive surface layer with pressure sensors \cite{schafer2023investigating}. This approach provides an alternative tracking method by detecting the pressure distribution when agents move over it, potentially enhancing the accuracy and reliability of localization through sensor data fusion.

The Kilogrid testbed utilizes a distributed array of modules, each capable of bidirectional infrared communication, to determine the positions of Kilobots based on their proximity to individual modules.

\subsubsection{\textbf{Accuracy}}
Both the IDS3C and the Robotarium, which utilize the Vicon system, report localization errors below \SI{1}{\milli \meter}. The Cambridge Minicar testbed, which uses the OptiTrack system, reports an even higher degree of accuracy, with localization errors below \SI{0.2}{\milli \meter}. These values indicate that global positioning systems based on high-precision cameras can achieve extremely high localization accuracy, providing a reliable foundation for complex navigation tasks.

The camera-based global positioning system in the \mbox{CPM Lab}, while not as precise as the Vicon or OptiTrack systems, still provides an acceptable level of accuracy for a significantly lower cost. With the reported localization error under \SI{3}{\centi \meter}, the \mbox{CPM Lab}'s system can adequately support a broad range of autonomous navigation tasks.

Conversely, for testbeds that primarily use agent-centric localization, such as Duckietown and F1TENTH, the localization accuracy can vary significantly based on the quality of the sensor data, the sophistication of the user's algorithm, and the algorithm's ability to handle uncertainties and noise.

\subsubsection{\textbf{Traffic Management}}
Duckietown and \mbox{CHARTOPOLIS} provide the incorporation of traffic management systems, which include traffic signs and traffic lights. These systems regulate and manage agent movement within the testbed, thereby creating dynamic and interactive environments. The inclusion of traffic management systems allows these testbeds to simulate real-world urban traffic conditions, providing a platform for testing and developing algorithms for traffic rule compliance, intersection management, and multi-agent coordination.

The F1TENTH, IDS3C, and \mbox{CPM Lab} testbeds take this concept further by introducing moving obstacles such as pedestrians into the environment. These obstacles present the agents with more unpredictable driving scenarios, necessitating more complex planning capabilities. The inclusion of moving obstacles thus presents a platform for researching and testing dynamic obstacle avoidance algorithms.

\subsubsection{\textbf{Surroundings}}
The IDS3C, Duckietown, and \mbox{CHARTOPOLIS} testbeds present structured environments that mimic real-world settings. The surroundings in these testbeds consist of scenery (e.g., trees, grass) and buildings, adding layers of complexity and realism to the navigational challenges. These additions allow the testbeds to simulate various urban scenarios, thereby enabling comprehensive testing of navigation algorithms under diverse and complex conditions. Besides physical surroundings, the ARK testbed provides virtual tasks such as foraging within its virtual environments, as demonstrated in \cite{font2018quality, talamali2020sophisticated} with hundreds of agents. In addition to foraging, the Kilogrid testbed demonstrates in \cite{valentini2018kilogrid} the virtual task of plant watering.

\subsection{Plan-based Information}
As we delve deeper, it is essential to understand the foundational architectures and computational frameworks that enable these testbeds to function effectively. This section begins by exploring the diverse architectures that underpin these testbeds.
\subsubsection{\textbf{Testbed Architecture}}
Multi-layered architectures have been widely used in small-scale testbeds, such as IDS3C, \mbox{CHARTOPOLIS}, Robotarium, and \mbox{CPM Lab}.
For example, the architecture of the \mbox{CPM Lab} features three layers (high-, mid-, and low-layer) and a middleware. The high-level layer is situated on the computational units, playing a vital role in planning. The mid-level and low-level layers, responsible for controlling the agents' actuators, are deployed on the agents. A unique aspect of this architecture is the usage of middleware that ensures deterministic and reproducible experiments. Communication is enabled through a DDS that operates via WLAN.

In contrast, Duckietown employs a unique communication strategy among its agents. Rather than using conventional wireless communication, it utilizes onboard LEDs that are detected by other agents. This method presents an alternative avenue for exploring agent communication strategies, demonstrating the adaptability and breadth of potential designs for testbed architecture.

The Kilogrid testbed consists of three parts: a set of Kilogrid modules, a dispatcher (interfacing a grid of connected modules with a remote workstation), and the KiloGUI application (allowing users to load and execute controllers both for Kilogrid modules and Kilobots). 
The ARK testbed also consists of three parts: a base control software, a modified overhead controller for communication, and several overhead cameras for tracking.

\subsubsection{\textbf{Distributed Computation}}
Most testbeds adopt a distributed approach to computational tasks. This methodology provides the advantage of computational load balancing, error resilience, and localized planning capabilities, enabling a higher degree of autonomy for individual agents.

The \mbox{CPM Lab}'s distributed approach is realized by incorporating an Intel NUC for each agent; notably, the Intel NUCs are not onboard, but rather serve as external computational units. This design effectively delegates computational tasks to each agent's dedicated unit, thereby facilitating parallel processing and real-time planning. Similarly, Duckietown, F1TENTH, \mbox{CHARTOPOLIS}, and Kilogrid use a distributed computation approach, where each agent relies on its own computational power. This allows for greater autonomy and real-time planning on a per-agent basis, enhancing the system's capacity to respond to dynamic changes in the environment.

In contrast, testbeds such as IDS3C that do not utilize a distributed computation approach may centralize their computations, which can offer a different set of advantages such as global system coherence and simplified data management. 

\subsubsection{\textbf{Computation Unit}}
In the \mbox{CPM Lab}, the Intel NUC computation units are located offboard to keep the weights of the agents low. In contrast, F1TENTH operates at a larger scale (1/10), which allows it to carry more substantial hardware components. Specifically, it employs a Jetson TX2, an advanced computation unit that provides significant computational power, integrated directly onto the agent. Similarly, \mbox{CHARTOPOLIS} uses a modified JetRacer Pro AI robotic car as the agent, which employs the Jetson Nano as a computation unit.
Further, while the Kilogrid mainly uses onboard computation, the ARK testbed additionally incorporates a GPU-enhanced base station to process data and control large swarms in a centralized manner.
In the Robotarium, a central server executes code and sends velocity control inputs to agents for onboard velocity tracking.
The choice of hardware components for these testbeds reflects the careful balance between the scale of the agents, the computational requirements of the tasks, and the physical constraints imposed by the hardware.

\subsubsection{\textbf{Computation Schemes}}
The \mbox{CPM Lab} provides support for a range of computation schemes, allowing parallel, sequential, and hybrid computations. This flexibility accommodates various task requirements and optimizes system performance by enabling the simultaneous processing of tasks or sequential execution as per the task dependencies \cite{kloock2023architecture}. Duckietown also accommodates a variety of computation schemes through its API, supporting both parallel and sequential computations. Moreover, it facilitates both onboard and offboard computation, further broadening the scope of its computation strategies. An edge computing station in \mbox{CHARTOPOLIS} is used to extend the testbed's computational capabilities beyond the computation onboard the agents when needed for tasks such as platooning and Vehicle-to-Infrastructure (V2I) communication. F1TENTH and Kilogrid primarily focus on onboard computation, potentially constraining their computation schemes to those suited for real-time, agent-based processing.
ARK supports both centralized and distributed computation \cite{reina2017ARK}. 

\subsubsection{\textbf{Human-robot Interaction}}
The IDS3C, Cambridge Minicar, \mbox{CHARTOPOLIS}, and \mbox{CPM Lab} testbeds offer capabilities to incorporate humans in the loop. Human interaction with these platforms is implemented in various ways, each with its own set of limitations. The specific methodologies and constraints depend on factors such as the overall system design, the complexity of tasks, and the degree of human involvement required. For example, the \mbox{CPM Lab} \cite{scheffe2023scaled} and CHARTOPOLIS offer a human-driven vehicle controlled manually with a steering wheel and pedals while a camera streams first-person-view images from the vehicle.

\subsubsection{\textbf{Different Kinds of Agents}}
While some testbeds like \mbox{CPM Lab} use homogenous agents, others support heterogeneous agents. 
For example, the Robotarium supports different agents with similar specifications, and Duckietown and IDS3C both include drones in their testbed environments, thereby introducing an aerial dimension. 
Notably, ARK has been recently extended to a new testbed called Multi-ARK (M-ARK) to enable the study of multiple, possibly heterogeneous \acp{rs}, e.g., aerial swarms interacting with ground swarms \cite{feola2023multi}.

\subsubsection{\textbf{Agent Count}}
The \mbox{CPM Lab}, IDS3C, Robotarium, and Cambridge Minicar testbeds offer up to 20 agents for experimentation. This sizable fleet enables the exploration of complex interactions and emergent behaviors in multi-agent scenarios.
In contrast, the agent count in Duckietown, F1TENTH, Kilogrid, and ARK can be varied based on the specific scenario. The flexible count in these testbeds allows for a customizable experimental setup, offering scalability based on the demands of the study being conducted.

\subsection{Act-based Information}
In this section, we explore the act-based characteristics of the testbeds, including agent dynamics and power consumption. These factors are crucial in determining how effectively agents can identify actions that align with planned trajectories and execute them.
\subsubsection{\textbf{Dynamics}}
The agents in the testbeds surveyed in this article are either differentially-driven, such as the Duckiebots (Duckietown) and the GRITSBots (Robotarium), or have an Ackermann steering geometry, such as the \textmu{}Cars (\mbox{CPM Lab}) and the F1TENTH agents.
For agents with Ackermann steering geometry, there are many common models which differ in accuracy and complexity.
In this overview, we consider the parameters for the point-mass model and the kinematic single-track model, which is based on geometric parameters of an agent. It captures the nonlinearity of the agent's motion while being simple to parameterize.
The parameters for the point-mass model, the nonholonomic differential-drive model, and the kinematic single-track model of the presented testbeds are listed in the online table \cite{webpage}.

Opting for low cost, the Kilobot agents (Kilogrid and ARK) use two sealed coin-shaped vibration motors for locomotion.
Since the motors are independently controllable, the agents move in a differential-drive manner.

\subsubsection{\textbf{Geometry}}
The geometry of an agent determines the occupied area around its center of gravity and directly impacts its ability to navigate and complete assigned tasks. Each testbed’s choice of geometry aligns closely with its objectives, balancing factors such as maneuverability, stability, and spatial requirements. Among the \mainTestbeds testbeds, there is significant variation in agent dimensions and weights, reflecting the diversity of applications these testbeds serve. For example, agents in the ARK and Kilogrid testbeds, which are tailored for swarm robotics studies, are among the smallest and lightest, with a wheelbase of \SI{33}{\milli \meter} and a weight of mere \SI{16}{\gram}. In contrast, the F1TENTH testbed, specifically designed for high-speed autonomous racing, features agents weighing over \SI{3000}{\gram}, prioritizing performance over compactness.

\subsubsection{\textbf{Battery}}
Unsurprisingly, due to the convenience of electrical batteries, the agents of all testbeds investigated in this article use them as an energy source.
Among these \mainTestbeds testbeds, the agents of the Kilogrid and ARK testbeds have the longest runtime---three to ten hours in the active mode and three months in the sleep mode---while maintaining a low ratio of battery weight to agent weight.
The agents in the \mbox{CPM Lab} also achieve a substantial runtime of up to five hours with a battery-to-agent weight ratio of one third.
Importantly, the battery selection for a testbed  depends on the intended objective of the testbed. For example, testbeds for autonomous racing like F1TENTH usually prioritize the battery-to-agent weight ratio over runtime. The battery runtime in F1TENTH is typically less than one hour, which is sufficient to finish a competition, while the battery-to-agent weight ratio is usually less than 20\%.
The agents in some testbeds may have a lower battery runtime, but with a simplified charging process. For instance, although the agents in the Robotarium only have a battery runtime of 40 minutes, they can automatically move to charging stations when their battery level is low, without human intervention.

\section{Additional Considerations}\label{sec:discussion}
Recall that the derived \characteristicsCount characteristics from the sense-plan-act paradigm are primarily objective, aiming to offer unbiased information on the existing small-scale testbeds.
However, during our investigation, we identified several aspects not covered by these characteristics.
For example, the agent dynamics in small-scale testbeds often lack the realism of full-scale deployments, which is hard to describe with an objective characteristic.
Another aspect that is difficult to quantify objectively is the sustainability of testbed development and operation.
Many of these missing aspects are covered by three ongoing challenges that we have identified: small-scale to full-scale transition (Section~\ref{subsec:smallToFull}), sustainability (Section~\ref{subsec:sustainability}), and power and resource management (Section~\ref{subsec:power}).
We will discuss these challenges in this section, illustrated with concrete examples from the selected testbeds. In addition, in Section~\ref{sec:discussion:communication}, we will explore another aspect overlooked by the sense-plan-act paradigm: communication.

\subsection{Small-Scale to Full-Scale Transition}
\label{subsec:smallToFull}
Transitioning from small-scale testbeds to full-scale real-world deployments presents several significant challenges that need to be addressed to ensure the successful scaling of technologies developed in controlled environments. Here, we explore these challenges in depth, discussing potential strategies for overcoming them and ensuring robust, scalable solutions.
\subsubsection{\textbf{Localization Challenges}}
Small-scale testbeds are typically situated indoors, and localization algorithms sometimes rely on global positioning systems that may not be directly transferable to outdoor, full-scale environments. Examples are the indoor position systems used in the \mbox{CPM Lab} \cite{kloock2020vision} and the overhead camera system used in ARK.
\subsubsection{\textbf{Sensing System Adaptation}}
Sensing algorithms developed in small-scale testbeds benefit from controlled lighting conditions and often depend on clear visual markers or well-defined road markings. In real-world deployments, sensing systems must adapt to diverse, unpredictable lighting conditions and handle scenarios where road markings are obscured by wear or environmental conditions such as snow or rain. Developing robust sensing systems requires the integration of more adaptive, context-aware algorithms capable of functioning reliably under varied environmental conditions.
\subsubsection{\textbf{Dynamic and Uncertainties Modeling}}
Small-scale testbeds often use agents with dynamics that do not fully capture the complexities encountered in full-scale deployments, such as varied payloads, tire wear, and suspension settings. This discrepancy raises questions about whether the planning and control algorithms developed in small-scale testbeds can be seamlessly applied in full-scale experiments. Additionally, full-scale agents must contend with environmental uncertainties such as road gradients, diverse road textures, and adverse weather conditions that can impact driving behavior, including hydroplaning or decreased stability on uneven roads. 
Moreover, uncertain events like traffic incidents, police pullovers, and breakage are rarely considered in small-scale testbeds.
Enhancing agent dynamics and incorporating real-world uncertainties to accurately reflect these complexities may be crucial for the successful translation of driving technologies from small-scale to full-scale. Although testbeds for autonomous racing, such as the F1TENTH testbed, address some aspects of this challenge, significant research is still required. 
\subsubsection{\textbf{Integration of Diverse Traffic Components}}
All of the small-scale testbeds we have investigated lack the diversity of real-world traffic environments that include elements like human-driven vehicles, pedestrians, and cyclists. Although some testbeds like IDS3C, the Cambridge Minicar, CHARTOPOLIS, and the \mbox{CPM Lab} model human-robot interaction, they suffer from limited fidelity compared with such interactions in the real world. Moreover, scaling up requires the integration of more complex dynamic models of traffic behavior that capture the full spectrum of interactions and the unpredictable nature of human elements in the traffic system. 

We also refer interested readers to a study of small-scale to full-scale transition in the \ac{cav} domain \cite{schafer2024smallscalea}, which discusses the transition challenge in terms of four aspects: vehicle, communication, automation, and environment. 
Moreover, CHARTOPOLIS testbed studies bridging the simulation-to-reality gap using domain adaption and domain randomization techniques in \cite{ulhas2024gan}.

\subsection{Sustainability}
\label{subsec:sustainability}
Building sustainable testbeds for \domainName{} is crucial, yet most testbeds are primarily focused on algorithm development without a targeted approach towards sustainability. The sustainability of small-scale testbeds concerns the capacity to develop and operate them in ways that are economically viable, environmentally sound, and socially responsible.
We will discuss how to increase testbed sustainability from both hardware and software perspectives.
\subsubsection{\textbf{Hardware Sustainability}}
A comprehensive needs assessment is critical. For instance, determining the scale of the testbed should align with its intended scope. For example, using or developing advanced sensing algorithms may demand a larger scale to accommodate extensive sensor arrays, as illustrated by the larger scale used in F1TENTH (1/10) than in \mbox{CPM Lab} (1/18). In addition, choosing a modular design is vital for enhancing the lifetime and adaptability of testbeds. This allows components to be easily upgraded or replaced, which reduces waste and enables the integration of new technologies without requiring complete system overhauls. Additionally, using rechargeable and swappable batteries can significantly increase the lifetime of the agents. Further, regular maintenance extends the lifetime of hardware components and ensures that they operate efficiently. 

In addition, integrating real and simulated environments through mixed reality, as demonstrated in the ARK testbed \cite{reina2017ARK}, reduces dependence on physical hardware, optimizing resource usage and easing maintenance demands. Furthermore, the M-ARK testbed \cite{feola2023multi}, an extension of ARK, networks multiple testbeds across different locations, which holds great potential to promote distributed resource utilization and remote collaboration. This mitigates the environmental impacts associated with transportation (i.e., traveling to other labs) and physical resource duplication.
\subsubsection{\textbf{Software Sustainability}}
Using open standards and open-source software reduces costs, promotes interoperability, and fosters innovation through community collaboration. This approach not only enhances the sustainability of the software itself, but also encourages global participation and knowledge sharing. 

Moreover, using effective power management measures, which we discuss next, can also contribute to a testbed's sustainability by increasing the lifetime of the agents.

\subsection{Power and Resource Management}
\label{subsec:power}
Small-scale testbeds often incorporate agents that have limited onboard energy capacity. Efficient management of power is essential to maximize the agents' operational time.
This is particularly important for testbeds that are used very frequently, where recharging or replacing batteries often is impractical.
In addition to power limitations, small-scale testbeds may face constraints in computational and communication resources. Effective management ensures that these resources are reasonably allocated and used.
In addition, by optimizing the use of available resources, the need for additional or more powerful hardware can be reduced, thus lowering the overall hardware expense for building a testbed.

Simple battery protection measures are used in some testbeds. 
For instance, the \mbox{CPM Lab} testbed prevents agents' batteries from over-discharging, 
and F1TENTH uses a power management board to ensure stable voltage for the computation unit and peripherals of the agent. The Robotarium testbed employs an efficient power management system and Qi inductive chargers that allow agents to charge autonomously. This setup minimizes human intervention and maximizes operational uptime, reflecting a sustainable approach to power management in small-scale testbeds. Additionally, it ensures that agents are always ready for new tasks, which is vital for high-throughput research and educational environments. For safety, wireless inductive chargers should be preferred over conductive rail charging, as the latter may encounter short circuits, which may lead to serious accidents \cite{wilson2020robotarium}. This is particularly relevant to testbeds that are designed to be frequently used (e.g., educational testbeds), where conductive charging has the potential to harm visitors who may accidentally bridge the charging leads \cite{wilson2020robotarium}. 

Although battery protection measures extend a battery's lifetime, aging occurs continuously. For example, although the \mbox{CPM Lab} employs a battery protection measure, the battery runtime drops more than 50\% in three years. Therefore, when building a new testbed, we recommend designing batteries to be swappable (rather than fixed-mounted), which would also increase the sustainability of the testbed. 

Additionally, it is essential to implement resource management measures to make full use of the limited power resource. For example, the Duckietown testbed employs two fundamental strategies for efficient usage of its resources. The first one is event-based computation, which uses an event-driven architecture to reduce the computational load by processing only necessary data changes rather than running continuous checks, thereby conserving computational resources and power. The second one is mode-driven sensing, which further optimizes resource usage: sensing tasks are selectively activated depending on the agent's current mode, which again ensures that the computational resources are used only when necessary. Another example is the Kilobot agents' use of a power control scheme to manage their operational states with a low-power sleep mode where the battery remains connected, instead of using a physical on/off switch. In this mode, the agent minimizes power consumption and periodically checks for a wake-up message from an overhead controller. This approach allows the agent to stay on standby for over three months on a single charge. Agents can display their battery status using colored LEDs, which helps operators assess the charge level and manage operations effectively. 

\subsection{Communication}\label{sec:discussion:communication}
Communication is vital for \acp{cav} and \acp{rs}, as they may heavily rely on communication for coordination.
In dynamic environments, the ability to share and receive data in real time allows agents to react to changes in their environment effectively.
This includes avoiding collisions with obstacles, cooperating with other agents, or responding to system instructions. 
Effective communication protocols are vital for scalability. As the number of agents in the testbed increases, the communication framework must efficiently handle the increased data traffic without loss of performance or increased latency.
For example, the ARK testbed modifies the traditional overhead controller system by implementing an enhanced infrared communication design, using a network of mini-overhead controllers for efficient, scalable messaging in large \acp{rs}.
It provides complete communication coverage of the testbed and uses additional lighting to mitigate the problem of shadows, thus decreasing communication uncertainties.
The Kilogrid testbed further enhances communication capabilities by including a bidirectional communication channel between agents and a remote workstation through a grid of computing nodes. 
Each grid module can independently communicate with agents via infrared messages, enabling decentralized, more scalable communication. 
Consequently, Kilogrid's communication bandwidth is 15 times higher than ARK's (144 bit/s versus 9.6 bit/s) \cite{valentini2018kilogrid}.

In testbeds where components for planning are not onboard the agents, such as the Robotarium and \mbox{CPM Lab}, these components must communicate plans to the agents, raising concerns about communication delays. For instance, \mbox{CPM Lab} uses Intel NUCs as components for planning, which utilize WLAN for communication. The effect of communication delays is mitigated by using a modular, hierarchical architecture with synchronized network components and a logical execution time approach, as described in \cite{kloock2023architecture}.

In \cite{cianci2007communication}, the authors presented a custom module for local radio communication designed for the e-Puck agents \cite{mondada2009puck}. This module allows the agents to exchange information locally with a software-controlled adjustable transmission power to manage effective communication ranges. This capability limits communication in confined spaces and enables the demonstration of collective behaviors of \acp{rs} based on local information, effectively avoiding the network saturation that may occur with broader-range radios. Instead of using infrared communication or radio-based communication, the agents in the Robotarium testbed communicate via WiFi with a bandwidth of up to 54 Mbit/s, leading to a theoretical upper limit on the number of agents of 18,000, given a typical bandwidth requirement of 3 kB/s per agent. Although uncertainties such as WiFi collisions will impair the agents' bandwidth, operating hundreds of agents in practice is feasible \cite{pickem2017robotarium}.

While different communication hardware and protocols have their strengths and limitations, the selection depends on the specific requirements of the testbed, such as the number of agents, the environmental conditions, and the required communication bandwidth.

\subsection{Limitations of Our Study}
Our online table \cite{webpage} includes \TestbedCount testbeds and \SingleAgentCount agents, featuring \characteristicsCount derived characteristics based on the sense-plan-act paradigm.
However, this paradigm has several limitations.
First, it is unsuitable for describing testbeds that employ end-to-end methods, where neural networks integrate the traditional sense-plan-act steps into a single process.
Second, this paradigm does not adequately cover communication, as it mainly focuses on the domains of sensing, planning, and acting. We have mitigated this shortcoming by discussing communication in Section \ref{sec:discussion:communication}.
Third, the online table does not encompass all the characteristics of testbeds. Each testbed may have unique characteristics that are not applicable to others. Including all these characteristics would result in an overly extensive table with many ``Not applicable'' or ``Information not found'' entries. Therefore, we balanced commonality and coverage when deriving the characteristics. 

Additionally, beyond the three challenges that we discussed in this section, there are other ongoing challenges with developing small-scale testbeds for \domainName{} that are out of the scope of this article, such as 
cybersecurity \cite{pascale2021cybersecurity} and ethical issues \cite{ulhas_chartopolis_2022, lin2014robot}.

\section{Conclusions}
This survey provided a detailed overview of small-scale \domainName{} testbeds, with the aim of helping researchers in these fields to select or build the most suitable testbed for their experiments and to identify potential research focus areas. We structured the survey according to characteristics derived from potential use cases and research topics within the sense-plan-act paradigm. Through an extensive investigation of \TestbedCount testbeds, we evaluated  \characteristicsCount characteristics and made the results of this analysis available in our online table \cite{webpage}.
We invited the testbed creators to assist in the initial process of gathering information and updating the content of this online table. This collaborative approach ensures that the survey maintains its relevance and remains up to date with the latest developments. The ongoing maintenance allows researchers to access the most recent information.

In addition, this article can serve as a guide for those interested in creating a new testbed. The characteristics and overview of the testbeds presented in this survey can help identify potential gaps and areas for improvement. 
As supplements, we discussed three ongoing challenges that we identified with small-scale testbeds, i.e., small-scale to full-scale transition, sustainability, and power and resource management.
Overall, this article provides a resource for researchers and developers in the fields of connected and automated vehicles and robot swarms, enabling them to make informed decisions when selecting or constructing a testbed and supporting the advancement of testbed technologies by identifying ongoing challenges.

\bibliographystyle{IEEEtran}
\bibliography{zoteroReferences,references}

\end{document}